\pdfoutput=1

\documentclass[11pt]{article}

\usepackage[]{EMNLP2022}
\usepackage{times}
\usepackage{latexsym}
\usepackage[T1]{fontenc}
\usepackage[utf8]{inputenc}
\usepackage{microtype}

\usepackage[normalem]{ulem}

\usepackage{soul}
\usepackage{url}
\usepackage[hidelinks]{}
\usepackage[utf8]{inputenc}
\usepackage{caption}
\usepackage{graphicx}
\usepackage{amsmath}
\usepackage{booktabs}
\usepackage{tikz}
\tikzset{every picture/.style={remember picture}}
\usetikzlibrary{tikzmark}
\usepackage{multirow}
\usepackage{makecell}
\usepackage{hhline}
\usepackage{paralist}

\usepackage{xcolor}
\newcommand\sect[1]{\S\ref{#1}}

\usepackage{graphicx,calc}
\newlength\myheight
\newlength\mydepth
\settototalheight\myheight{Xygp}
\newcommand*\inlinegraphics[1]{%
  \settototalheight\myheight{Xygp}%
  \settodepth\mydepth{Xygp}%
  \raisebox{-\mydepth}{\includegraphics[height=\myheight]{#1}}%
}

\usepackage{enumitem}
\usepackage{amssymb,amsmath,amsthm,enumitem}

\title{Inferring the Reader: Guiding Automated Story Generation with Commonsense Reasoning}

\author{Xiangyu Peng\textsuperscript{$\clubsuit$}\Thanks{\ Equal contributions} \hspace*{10mm} Siyan Li\textsuperscript{$\clubsuit$}\footnotemark[1] \hspace*{10mm} Sarah Wiegreffe\textsuperscript{$\dagger$}  \hspace*{10mm}  Mark Riedl\textsuperscript{$\clubsuit$}\\
\textsuperscript{$\clubsuit$}Georgia Institute of Technology\\
\textsuperscript{$\dagger$}Allen Institute for Artificial Intelligence \\
\texttt{\{xpeng62, sli613\}@gatech.edu}\\
\texttt{wiegreffesarah@gmail.com}\\
\texttt{riedl@cc.gatech.edu}}

\begin{document}
\maketitle
\begin{abstract}
Transformer-based language model approaches to automated story generation currently provide state-of-the-art results.
However, they still suffer from plot incoherence when generating narratives over time, and critically lack basic commonsense reasoning.
Furthermore, existing methods generally focus only on single-character stories, or fail to track characters at all.
To improve the coherence of generated narratives and to expand the scope of character-centric narrative generation, we introduce Commonsense-inference Augmented neural StoryTelling (CAST), \footnote{Code: \url{https://github.com/xiangyu-peng/CAST_public}} a framework for introducing commonsense reasoning into the generation process with the option to model the interaction between multiple characters. 
We find that our CAST method produces significantly more coherent, on-topic,  enjoyable and fluent stories than existing models in both the single-character and two-character settings in three storytelling domains.
\end{abstract}

\section{Introduction}
AI storytelling is a crucial component of computational creativity. 
Humans use storytelling to entertain, share experiences, educate, and to facilitate social bonding \cite{riedl:jair2010}.
For an intelligent system to be unable to generate a story limits its ability to interact with humans in naturalistic ways~\cite{riedl:chi-hcml2016}.
Automated Story Generation, the task of requiring a system to construct a sequence of sentences that can be read and understood as a story, is a grand challenge in AI.
\begin{figure}[t]
    \includegraphics[width=\columnwidth]{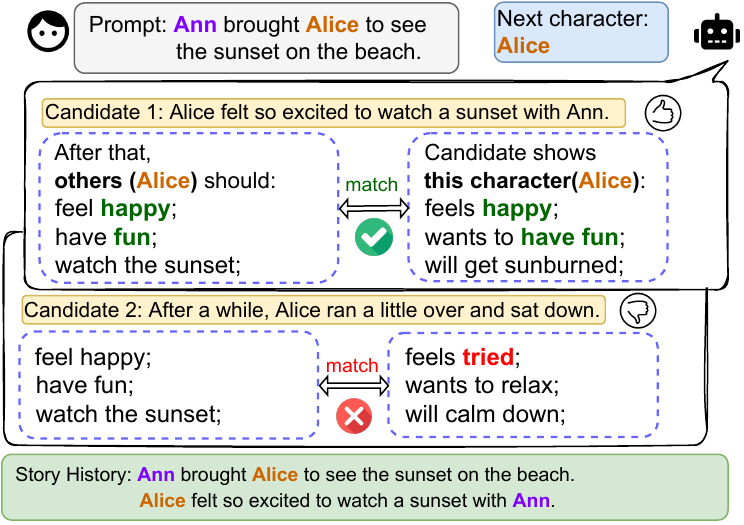}
    \caption{Overview of the CAST system \inlinegraphics{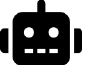}. 
    1. A text prompt and a specified character start the story generation process. 
    2. A language model generates candidate continuations (two are shown) with the specified character as the main character. 
    3. The system infers commensense attributes about the main character from each candidate sentence.
    4. If enough inferences from a candidate sentence match those from the prompt sentence, the candidate is added to the story and becomes the new prompt (here, only the first candidate meets this criterion). 5. The process repeats (with the option to specify a new main character) until a story of desired length is generated.
    }
    \label{fig:overview}
\end{figure}

Prior to the advent of neural language models, methods to model the narrative arcs of stories leveraged a variety of statistical techniques to track events and characters \cite{gervas2013propp, gervas2014composing, ouyang-mckeown-2015-modeling}. 
The dominant approach to story generation today is to use neural language models~\cite{roemmele2016writing,khalifa2017deeptingle,clark2018,martin2018event}. When a language model is trained on a corpus of stories, samples from the resulting distribution tend to also be stories.
These techniques have improved with the adoption of Transformer-based models, such as GPT-2~\cite{radford2019language} and GPT-3~\cite{brown2020language}. However, these models are prone to generating repetitive or generic continuations~\cite{holtzman2019curious}. 
Furthermore, as the length of the story grows, these models can lose coherence. Other artifacts include new characters being arbitrarily introduced at any time and characters being forgotten.
One reason for these phenomena is that language models generate continuations by sampling from a learned distribution $P_\theta(tok_n | tok_{<n})$.
Human readers, however, do not perceive the coherence of a narrative as a function of the likelihood of seeing specific continuations of previous contexts. Statistical sampling from a distribution is not constrained to making logical transitions because the rich relationships that readers make to perceive coherence are not modeled.

Previous attempts to enhance story generation coherence use conditioning on content-relevant features such as plot outlines~\cite{fan2018hierarchical,peng-etal-2018-towards,rashkin2020plotmachines}, or character emotional arcs~\cite{brahman-chaturvedi-2020-modeling}.
These improve plot coherence through adherence to a manually-given high-level plan.
A high level plan can also be automatically generated then decomposed~\cite{yao2019plan,fan2019strategies,ammanabrolu2020story}, which elevates the challenges of maintaining coherence to a higher level of abstraction.  
Neural language models can also be fine-tuned on other signals such as commonsense knowledge or progression rewards~\cite{guan2020knowledge,tambwekar2018controllable}, which improves the distribution but still relies solely on sampling and the assumption that language models can encode complex story structure in token distributions.

The latent state of neural language models used to generate subsequent story continuations are unlikely to relate to a human reader's mental model of the state of a story world.
Studies of human reader comprehension~\cite{trabasso1985causal,graesser91,graesser94} show that readers comprehend stories by tracking the relations between events. 
Specifically, reader comprehension relies on the tracking of at least four types of relations between events: (1)~causal consequence, (2)~goal hierarchies, (3)~goal initiation, and (4)~character intentions. 
The perceived coherence of a story is thus a function of the reader being able to comprehend how events correlate to each other causally or how they follow characters' pursuits of implicit goals.
We hypothesize that a story generation system that makes decisions on how to continue a story based on tracking and reasoning about events will generate more coherent stories.

Unfortunately, stories don't always explicitly declare the causal consequences of events or the goals and intentions of characters. 
That is, sentences describing character actions or external events are rarely explicitly annotated with the characters' motivations and goals. 
Readers must infer the characters' goals, the relationship between their actions and those goals, and how their goals change as a result of the events in their world.
The ability to use basic knowledge about goals and about world states falls within the study of {\em commonsense inference}. Initial work in this area was limited to modeling dimensions of the ``naive psychology'' of characters: motivations and emotional reactions \cite{rashkin-etal-2018-modeling}.
This was later extended to more attributes---ATOMIC \cite{sap2018atomic} and ATOMIC$_{20}^{20}$ \cite{Hwang2021COMETATOMIC2O} are event-centric commonsense knowledge bases that contain logical relationships between events and the mental states and attributes of their participants, represented as typed \textit{if-then} relations. The former contains 9 such dimensions, and the latter 23.
COMET$_{20}^{20}$ \cite{Hwang2021COMETATOMIC2O}, an extension of COMET \cite{bosselut2019comet}, is a transformer-based generative model trained on triples from ATOMIC$_{20}^{20}$. Given a sentence, COMET$_{20}^{20}$ infers commonsense attributes about the characters that fall into three categories:     
    (1) social interactions,
    (2) physical entities, and
    (3) effect of events inferred from the sentence.
We hypothesize that a neural language generator informed about COMET-inferred event effects as well as character intentions and goals can generate more coherent narratives. 

To this end, we introduce {\em \textbf{C}ommonsense inference \textbf{A}ugmented neural \textbf{S}tory\textbf{T}elling} (CAST), which infers the causal relations between events as well as the intents and motivations of characters in the story so far in order to generate story continuations that are more coherent to readers.
CAST is a straightforward, cognitively inspired method to scaffold the generation of story text when sampling from a language model.
By chaining sentence-level COMET inferences to track important implicit elements of the story over time, CAST is able to make more informed choices when sampling story continuations from a neural language model of choice (GPT-2 in our experiments). It can be used to produce both single-character and multiple-character stories.
We hypothesize that stricter, more explicit constraints during generation should result in more coherent narratives than generating via sampling from a distribution alone, even if the distribution is fine-tuned. An overview of our method is presented in \autoref{fig:overview}. 

To evaluate the efficacy of our proposed method, we conduct a series of human-participant experiments that measure perceptions of logical coherence of CAST against three strong neural language model story generators on three different story corpora.
Results indicate that the CAST method produces significantly more coherent, on-topic,  enjoyable and fluent stories in both the single-character and two-character settings. This result holds even in a genre with a very different type of commonsense than that which COMET is trained on (fairy tales), indicating our method's generality.

\section{Related Work}

\begin{figure*}[t]
    \centering
    \includegraphics[width=\textwidth]{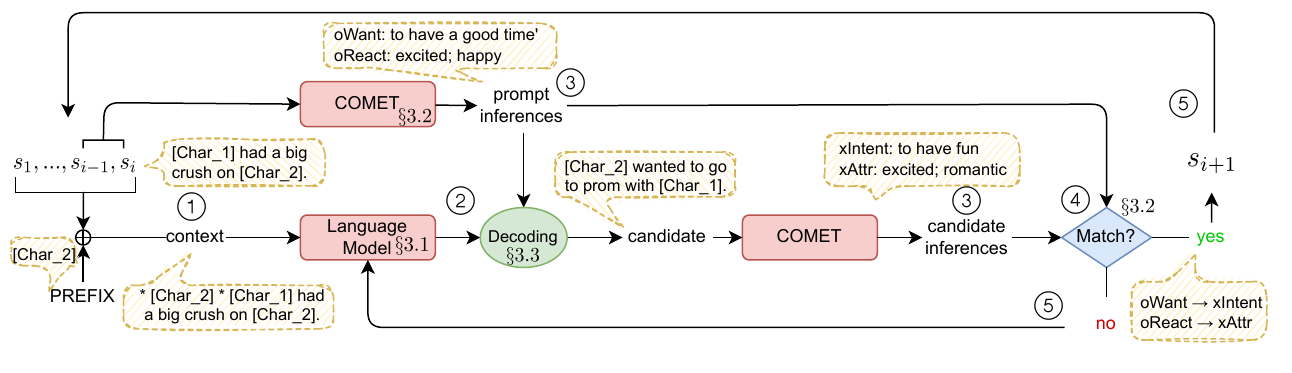}
    \caption{The overall procedure of generating two-character narratives with the CAST pipeline at step $t$.
    }
    \label{fig:pipeline}
\end{figure*}

In addition to the work mentioned in the introduction, we provide a detailed background on story generation systems that emphasize commonsense reasoning and other related techniques.
\citet{guan2019story} were the first to propose to incorporate a commonsense knowledge base into the story generation pipeline. \citet{guan2020knowledge} improved upon this method by using the ATOMIC dataset to fine-tune GPT-2, and then fine-tuning a second time on the ROCStories corpus~\cite{mostafazadeh2016corpus}. 
This system used multi-task learning during a second fine-tuning stage with an auxiliary objective to distinguish true and engineered false stories. 

Similarly, \citet{paul-frank-2021-coins} finetune GPT-2 on ROCStories to obey coherence rules generated by separately trained models. At inference-time, the story generation model is fed the first two and the last sentences of ROCStories test instances, making this an infilling rather than open-ended generation task.
\citet{brahman-chaturvedi-2020-modeling} fine-tune GPT-2 to generate stories that follow a given emotional arc for a protagonist, using COMET to infer the protagonist's emotions as labels for their training dataset. 
They assume five emotions (anger, fear, joy, sadness, neutral) limited to two changes throughout the story, associated with the \emph{xReact} and \emph{oReact} inferences produced by COMET. We do not assume a fixed set of commonsense inference values, and we assume a character's state may change at each new sentence.

The C2PO system~\cite{ammanabrolu2020automated} uses COMET to generate successor and predecessor events instead of a language model, performing a bi-directional search from a given start event and a given end event.
C2PO assembles the narrative directly from the short, templated sentences produced by COMET.
It also assumes the end of the story is known in advance. Like \citet{brahman-chaturvedi-2020-modeling}, it focuses on only two dimensions of COMET and only works for a single character. Our work models interactions between multiple characters and takes advantage of a richer set of inferences that COMET and COMET$_{20}^{20}$ provide, better aligning with the four types of relations key to reader comprehension \cite{trabasso1985causal,graesser91,graesser94}.

Very recently, \citet{lin2022inferring} utilizes a BART-based commonsense inference model in conjunction with an event generation model to place event-related constraints on the story generation process. We acknowledge high-level similarity in approaches between our framework and this work, but we do not include this approach in our comparisons since it is concurrent.

Storytelling research focused on improving long-range cohesion is not limited to using commonsense resources---\citet{goldfarb2020content} perform high-level planning via plot outline generation using principles from Aristotle's \emph{Poetics}, then use a language model in fill in details. They demonstrate strong performance on the WritingPrompts \cite{fan2018hierarchical} dataset. While one of their models' purpose is to determine whether to reuse or introduce a new character, they do not explicitly model inter-character relationships or character attributes during generation. 
We compare to this baseline in our experiments.

\section{The CAST Inference Method}
We now introduce our neural storytelling framework, {\em \textbf{C}ommonsense inference \textbf{A}ugmented neural \textbf{S}tory\textbf{T}elling} (CAST), which scaffolds the conventional text generation process by imposing constraints on the  sampling process at inference time.

The conventional setup for $k$-sentence story generation starts with a given first sentence $s_1$, referred to as the prompt, and generates $k-1$ subsequent sentences conditioned on it.
CAST follows this convention, generating the $i$th sentence as follows:

\begin{enumerate}[noitemsep,topsep=0pt,leftmargin=*]
    \item We condition a fine-tuned language model on the story up to the current sentence $[s_1,\dots, s_{i-1}]$, followed by a token signifying the main character of sentence $i$ (\S\ref{sec:lm}).
    \item We obtain a set of commonsense inferences for $s_{i-1}$ (\sect{sec:comet}) and use them as constraints at the decoding stage of sampling a next-sentence candidate $c$ from the language model (\sect{sec:decode}).
    \item We obtain a set of commonsense inferences for candidate $c$, and match commonsense inference sets between $s_{i-1}$ and $c$ using a matching criteria, producing a score for $c$  (\sect{sec:comet}).
    \item If the score is above a specified threshold, $c$ is selected to be $s_i$ and is appended to the generation history. Otherwise, steps 2 through 4 are repeated until a viable candidate is found.
    \item We repeat steps 1 through 4 until $k-1$ sentences have been generated.
\end{enumerate}

\noindent
An illustration of the pipeline is given in \autoref{fig:pipeline}. In practice, when generating two-character stories, we specify main characters in an alternating manner to promote turn-taking (more details in \sect{sec:lm}). CAST is not limited in application to the maximum story length seen during training, and can be used to generate stories of arbitrary length.

\subsection{Language Model}
\label{sec:lm}
We fine-tune GPT-2 on a story corpus to prime the model to elicit story-like generations (See Section~\ref{ssec:datasets} for details on the three story corpora we use for experiments).
In order to directly compare to prior work \cite{guan2020knowledge}, we use the small version of GPT-2, although our technique works with any neural language model.

We first pre-process the corpus to remove character names to improve generality and avoid gender bias.
We replace them with character tags such as \texttt{[Char\_1]}, \texttt{[Char\_2]} or \texttt{[Char\_3]}. This allows our generated stories to not be limited to turn-taking between characters of different genders, in contrast to prior work \cite{guan2020knowledge} who use gendered tags (\texttt{[MALE]}, \texttt{[FEMALE]}, \texttt{[NEUTRAL]}) to differentiate characters. 

We perform a second fine-tuning step where we append a special prefix \texttt{*$T$*} to sentence pairs. $T$ is a character tag (\texttt{[Char\_1]}, \texttt{[Char\_2]}, etc.) representing the character that is the subject of the second sentence. Fine-tuning on this corpus allows us to specify main characters during Step 1 of the CAST inference process using the \texttt{*$T$*} tag. This allows the model to generate a second sentence where $T$ is the subject, but not necessarily the first word, of the sentence. For example, consider the sentence ``\texttt{[Char\_1]} has a big crush on \texttt{[Char\_2]}''.
If the next sentence in the corpus has the entity represented by \texttt{[Char\_2]} as the subject of the event, then we concatenate \texttt{*[Char\_2]*} onto the first sentence during fine-tuning in order to cue the language model about turn-taking. We identify the subject entity in a sentence using a parser.
We found in initial experiments that this allowed more flexibility and improved generation quality over the alternative (always requiring the main character $T$ to be the first word of a sentence).
More details are given in Appendix \ref{app:appendix-prefix}.

To use the language model to generate a single-character story, we always set $T$ to the same character (\texttt{[Char\_1]}).
In a two-character story setting, we adopt character turn-taking principle: in a two-character 5-sentence story, $T = \texttt{[Char\_2]}$ for generating even numbered sentences and $T = \texttt{[Char\_1]}$ for generating odd-numbered sentences.

\subsection{Generating and Matching Commonsense Inferences}
\label{sec:comet}
To produce commonsense inferences for each sentence, we use the $\text{COMET}^{20}_{20}$ model~\cite{Hwang2021COMETATOMIC2O} to infer a set of commonsense relations for each prompt sentence $s_{i-1}$ and continuation sentence $c$.
Table~\ref{tab:def} has the list of the subset of inferences we use and their definitions. 

\begin{table}
\footnotesize
\centering
\setlength\tabcolsep{6pt}
\begin{tabular}{l|l}
\toprule
 Type & Definition\\
\midrule
\texttt{xWant} & as a result, \texttt{PersonX} wants \\
\texttt{xIntent} & because \texttt{PersonX} wanted\\
\texttt{xNeed} & before this event, \texttt{PersonX} needed\\
\texttt{xEffect} & as a result, \texttt{PersonX} will\\
\texttt{xAttr} & \texttt{PersonX} is seen as\\
\texttt{xReact} & as a result, \texttt{PersonX} feels\\
\texttt{oReact} & as a result, \texttt{PersonY} or others feels\\
\texttt{oWant} & as a result, \texttt{PersonY} or others want\\
\texttt{oEffect} & as a result, \texttt{PersonY} or others will\\
\texttt{CausesDesire} &	This event makes \texttt{PersonX} want\\
\texttt{Desires} &	\texttt{PersonX} desires\\
\bottomrule
\end{tabular}
\caption{Definitions of the selected types of ATOMIC$_{20}^{20}$ relations that CAST uses. Those prefaced with \texttt{x} refer to the sentence's subject character and those with \texttt{o} to other characters.
}
\label{tab:def}
\end{table}
\begin{table}[!tb]
\footnotesize
\centering
\setlength\tabcolsep{6pt}
\begin{tabular}{c|c|c}
\toprule
\# Characters & Prior Context & Continuation\\
\midrule
\multirow{6}{*}{Single} &
\texttt{xWant} & \texttt{xIntent}\\
&\texttt{xReact} & \texttt{xReact}\\
&\texttt{xEffect} & \texttt{xEffect}\\
&\texttt{xReact} & \texttt{xAttr}\\
&\texttt{CausesDesire} & \texttt{Desires}\\
\midrule
\multirow{3}{*}{Multiple} &
\texttt{oReact} &	\texttt{xAttr}\\
&\texttt{oWant} &\texttt{xIntent}\\
&\texttt{oEffect} & 	\texttt{xEffect}\\

\bottomrule
\end{tabular}
\caption{Commonsense relation pairs that we identify as leading to meaningful sentence continuations. 
Table~\ref{tab:def} provides attribute definitions.}
\label{tab:pairs}
\end{table}

Once we have inferred relation sets for both sentences, we look for specific patterns between the sets of ATOMIC relations. We identify eight relation pairs that are useful for creating coherent relations between story events described in adjacent sentences, five for single-character and three for two-character stories (Table~\ref{tab:pairs}) by analyzing semantic similarities between these relation pairs inferred by stories in ROCStories~\cite{mostafazadeh2016corpus}. 
Details on the process for finding relation pairs can be found in Appendix~\ref{app:pairs}.

The relation types in the second column of Table~\ref{tab:pairs} are interpreted as 
{\em postconditions} of a prior event because they are inferences of things that might have changed, such as an effect of the event, or a character is expected to form a new intention, or a character is expected to have a reaction.
The relation types in the third column are interpreted as
{\em preconditions} of any generated continuation because they are inferences about facts that needed to be established by prior events in the story, such as a character having an intention, a character having desires, or a character having a property.

CAST seeks to chain preconditions of the currently generated sentence with the post-conditions of the previous sentence as a means of checking whether readers will comprehend the continuation and perceive coherence.
For example, suppose sentence $s_{i-1}$ is ``\texttt{[Char\_1]} gives \texttt{[Char\_2]} a burger''.
From this one can infer an \texttt{oWant} is ``to thank'', indicating that \texttt{[Char\_2]} may want to thank \texttt{[Char\_1]}. If \texttt{[Char\_2]} is to be the subject of the subsequent sentence, a good candidate sentence would be one from which the \texttt{xIntent} ``to thank'' can be inferred, such as ``\texttt{[Char\_2]} said thanks to \texttt{[Char\_1]}''.

Once we have inferred relations for the previous sentence $s_{i-1}$ and a current candidate $c$, we judge the coherence of $c$ with the following procedure:

\begin{itemize}[noitemsep,topsep=2pt,itemsep=2pt]
    \item The event in $s_{i-1}$ affects the wants of a character, which manifests as an intention of the primary character in the subsequent sentence (\texttt{xWant}$\rightarrow$\texttt{xIntent} in a single-character story; \texttt{oWant}$\rightarrow$\texttt{xIntent} in a multi-character story). 
    
    \item An effect of the event in $s_{i-1}$ is something the primary character will do in the subsequent sentence (\texttt{xEffect}$\rightarrow$\texttt{xEffect} in a single-character story; \texttt{oEffect}$\rightarrow$\texttt{xEffect} in a multi-character story).
    
    \item A reaction to the event in $s_{i-1}$ should match either some property, or the reaction, of the primary character in the subsequent sentence (\texttt{xReact}$\rightarrow$\texttt{xAttr}/\texttt{xReact} in a single-character story and \texttt{oReact}$\rightarrow$\texttt{xAttr} in a multi-character story).
    \item The character's desire in $s_{i-1}$ should be consistent in the subsequent sentence (\texttt{CausesDesire}$\rightarrow$\texttt{Desires} in a  single-character story).
\end{itemize}

\noindent
To filter out ``unqualified" continuations generated by the language model, we match the inference types described in \autoref{tab:pairs} and their arguments.

In practice, we find that simple string matching does not adequately capture when two inferred relations' arguments have slightly different phrasing (e.g., ``to sleep'' versus  ``sleeping''). 
We define a match as the semantic similarity between two inferences exceeding a certain threshold. 
To do this, we encode each relation argument into a fixed-length vector representation, and then compute the cosine similarity. 
We use Sentence-BERT~\cite{reimers-2019-sentence-bert} for encoding, as it is designed for semantic similarity tasks and performs better than the traditional BERT on sentence similarity benchmarks.
We use $80\%$ semantic similarity as our threshold. 
In order to balance computation time and quality of the match, we require three of five and three out of three inference type pairs to match between a prompt and a candidate sentence when generating single-character and two-character stories, respectively. 
Details of ablation studies on these hyperparameters can be found in Appendix~\ref{app:ablation}.

\subsection{Increasing Sampling Success}
\label{sec:decode} 

CAST produces many candidates $c$---this step can be very expensive with respect to the average number of continuations needed to find a match (see Appendix~\ref{app:ablation}).
In order to increase the probability of generating a continuation with a match, we use the commonsense inference set of prompt sentences as lexical constraints to control the decoding process~\cite{peng2022xfboost} when generating candidates.
We first obtain the synonyms set $A$ and the antonyms set $\bar{A}$ of each commonsense inference output by COMET using WordNet \cite{miller1995wordnet}.
For example, if COMET infers a want for a character of ``go to beach'', then $A = {}$ \{``move to beach'',``go to beach''\} and $\bar{A} = $ \{``leave beach''\}. 
We then expand the synonyms and antonyms sets by adding conjugated forms of verbs and the plural and singular forms of nouns \cite{de2012pattern}.
For example, \{`` buy dog''\} is expanded to \{`` buy dogs'', ``buy a dog'' , ``buys a dog'', ``bought a dog'', ...\}.
Finally, we modify the conditional probability $P_\theta(x_{i} \mid x_{<i})$ of the language model prior to sampling as follows:
\begin{align*}
    P'_\theta(\vec{x_i} \mid x_{<i}) &=P_\theta(\vec{x_i} \mid x_{<i}, A, \bar{A}) \nonumber \\
    &= \delta(\vec{x_i}, A, \bar{A}) \times P_\theta(\vec{x_i} \mid x_{<i})
    \label{eq:modified_equation}
\end{align*}
where
\begin{equation*}
  \delta(\vec{x_{ij}}, A, \bar{A}) =
    \begin{cases}
      1 + \mu & \text{if $ x_{ij} \in A, x_{ij} \notin \bar{A}$}\\ 
      1 - \mu & \text{if $ x_{ij} \notin A, x_{ij} \in \bar{A}$}\\
      1 & \text{otherwise}
    \end{cases}       
\end{equation*}
and $\mu$ is a hyper-parameter to control the strength of the penalty. Using this altered distribution, we encourage productions of synonyms to COMET inferences and punish productions of antonyms. 
We only modify the probability of the \textit{top-k} tokens in order to maintain the fluency of generated sentences.
More details are in Appendix~\ref{app:decode}.

\section{Evaluation}\label{sec:experiment_setup}

\begin{table}[t]
\centering
\footnotesize
\setlength\tabcolsep{0.1pt}
\begin{tabular}{p{0.97\linewidth}}
\toprule
\textbf{Seed Prompt:}  \\
 Alice invited Megan and the girls over for a clambake. \\
\midrule
\textbf{CAST:}  \\
The rain was so bad that Megan \textbf{couldn't play }in the rain. \\
Alice got so \textbf{sad} and wanted to play in the rain with Megan.\\
Megan got a \textbf{rain coat }to go outside. \\
Alice finally \textbf{went outside} with Megan together.\\
\midrule

\textbf{GPT-ROC-RL:}\\
Alice and Megan decided to \textbf{\textit{stay inside}} for a week.\\
Alice and Megan \textbf{bathroom was broken}.\\ 
Alice and Megan got in so much \textbf{trouble}.\\ Alice and Megan decided to \textit{stay inside}.\\

\midrule

\textbf{\citeauthor{guan2020knowledge} \shortcite{guan2020knowledge}}\\
Megan tried to be friendly with each other.\\ Alice each could feel their \uwave{favorite animal tense} in their hands. \\
Megan caught up with them. \\
Alice \uwave{wrote down the activities} on a topic they would enjoy.\\
\bottomrule

\end{tabular}
\caption{Story examples generated by CAST, CAST-RL, GPT-ROC and \citeauthor{guan2020knowledge} \shortcite{guan2020knowledge}. The story generated by CAST follows a single topic (\textbf{bolded})---playing outside, and shows a good plot coherence. 
GPT-ROC-RL generates relatively more repetitive/boring but logically coherent narrative (in \textit{italic}). \citeauthor{guan2020knowledge} \shortcite{guan2020knowledge} suffers plot incoherence  (\uwave{underlined}). More examples are given in \autoref{app:extra_stories}.}
\label{tab:ex_story}
\end{table}

\begin{table*}[t]
\centering
\setlength\tabcolsep{0.2pt}
\resizebox{\textwidth}{!}{
\begin{tabular}{c|c|c|lll|lll|lll|lll}
    \toprule
    \multirow{2}{*}{\textbf{Models}} & \textbf{Data}&
    \textbf{Num} & \multicolumn{3}{c|}{\textbf{Logical Sense }} & \multicolumn{3}{c|}{\textbf{Single Topic }} & \multicolumn{3}{c|}{\textbf{Enjoyable}} & \multicolumn{3}{c}{\textbf{Fluency}}\\
    & \textbf{set} & \textbf{chars} & Win\% & Lose\% & Tie\% & Win\% & Lose\% & Tie\% &  Win\% & Lose\% & Tie\%  & Win\% & Lose\% & Tie\%\\
    \midrule

    \multirow{2}{*}{CAST vs \citeauthor{guan2020knowledge}} 
    &\multirow{2}{*}{ROC}
    & 1 & \textbf{92.0}**&4.0&4.0&  \textbf{86.0}**&7.0&7.0 & \textbf{87.0}**&4.0&9.0&\textbf{87.0}**&4.0&9.0\\
    
    && 2 &
    \textbf{85.8}**&6.6&7.5&
    \textbf{82.9}**&8.6&8.6&
    \textbf{81.1}**&12.3&6.6&
    \textbf{83.0}**&9.4&7.5\\
    
    \midrule
    CAST vs \citeauthor{goldfarb2020content}
    & WP
    & 1  &
    \textbf{64.2}*&32.1&3.8&
    \textbf{64.2}**&28.3&7.5&
    \textbf{62.3}**&26.4&11.3&
    \textbf{52.8}&34.0&13.2\\
    
    \midrule
    CAST vs C2PO
    &FT
    & 1 & 
     \textbf{81.5}**&9.3&9.3&
    \textbf{63.6}**&23.6&12.7&
    \textbf{81.8}**&10.9&7.3&
    \textbf{85.5}*&5.5&9.1\\
    \bottomrule
\end{tabular}
}
\caption{Human-participant evaluation results for experiments 1 and 2, 
showing the percentage of participants who preferred the first system, second system, or thought the systems were equal.
Each system is conditioned on the same test-set prompts. * indicates results are significant at $p<0.05$ confidence level; ** at $p<0.01$ using a Wilcoxan sign test on win-lose pairs. 
See results about majority votes and agreement in Table~\ref{tab:CAST_results_majority}. 
}
\label{tab:CAST_results}
\end{table*}

\subsection{Datasets}\label{ssec:datasets}
We conduct experiments on three diverse storytelling datasets:
\begin{itemize}[noitemsep,topsep=3pt,leftmargin=*]
    \item ROCStories (ROC) \cite{mostafazadeh2016corpus}: contains $98,159$ crowdsourced five-sentence stories involving common-sense scenarios. 
    \item Writing Prompts (WP) \cite{fan2018hierarchical}: \textasciitilde300K user-generated stories along with their associated prompts from Reddit (\texttt{r/WritingPrompts/}). Average story length is $59.35$ sentences.
    \item Fairy tales (FT) \cite{ammanabrolu2020bringing}: $695$ stories in the fairy tale genre scraped from story summaries on Wikipedia. Average length of stories is 24.80 sentences.
\end{itemize}

\subsection{Baselines} We evaluate CAST against three strong baselines. 
\begin{itemize}[noitemsep,topsep=0pt,leftmargin=*]
    \item \citet{guan2020knowledge}: fine-tunes GPT-2-Small on ATOMIC and ROCStories using a multi-objective training procedure. This baseline serves to demonstrate whether a neural language model can get everything it needs directly from a static commonsense dataset without inference and constraints. We retrain the model on the pre-processed version of the ROCStories corpus that does not contain gender tags (\sect{sec:lm}) as well as with the additional fine-tuning step for character-conditioned generation (\autoref{app:appendix-prefix}), in order to be directly comparable to CAST in a two-character setting. Further training details can be found in Appendix~\ref{app:baselines}.
    
    \item \citet{goldfarb2020content}:  a plot-generation language model along with an ensemble of rescoring models on Writing Prompts dataset.
    The system trained BART \cite{lewis2020bart} to learn to generate plots on the given prompt and then transform them into a story. 
    We compare CAST to \citet{goldfarb2020content}---one of the strongest story generators on Writing Prompts dataset---to show that CAST can be generalized to other datasets.
    
    \item C2PO \cite{ammanabrolu2020automated}: uses COMET to generate successor and predecessor events for a single character, performing a bi-directional search from a given start event and an end event.
    It uses COMET to generate successor and predecessor events directly instead of constraining a more conventional language model as is the case with CAST. As such it is a strong baseline, especially considering it uses an extra piece of input---the story ending---that can influence perceptions of coherence.
    For fair comparison, we follow \citet{ammanabrolu2020automated} to extract high-level plots from fairy tale stories and then use the first plot as prompt and the second plot as goal for guiding C2PO. 

\end{itemize}
We use the provided checkpoints of the latter two models.\footnote{\url{https://github.com/PlusLabNLP/story-gen-BART}; \url{https://github.com/rajammanabrolu/C2PO}} We thus only evaluate these systems on single-character stories, since C2PO is single-character story generator and \citet{goldfarb2020content} is not trained to generate stories with the number of characters chosen by humans.

\subsection{Metrics}
\label{sec:metrics}
Given the well-established unreliability of automated metrics\footnote{Perplexity and BLEU \cite{papineni2002bleu} scores are not applicable to evaluate CAST, because CAST is unconstrained neural language model story generator. It is not required to produce the same story with the gold story in the datsets. Self-BLEU~\cite{zhu2018texygen} measures frequency of words and bi-grams, which does not necessarily entail plot-level repetition; the same entities can make appearances in different events in different ways.} for creative text generation, human-participant evaluation is generally held as the gold-standard evaluation technique~\cite{celikyilmaz2020evaluation, caglayan-etal-2020-curious, van2021human}.
Consequently, we also use human-participant evaluation. 
We provide human participants with a pair of stories from two systems, and ask them the following questions modified from \citet{purdy2018predicting}:
\begin{itemize}[noitemsep,topsep=0pt,leftmargin=*]
    \item Which story better FOLLOWS A SINGLE TOPIC: for insight into perceptions of global coherence for the entire story.
    \item Which story's sentences MAKE MORE SENSE given sentences before and after them: to evaluate local causality and commonsense reasoning in the story.
    \item Which story is more ENJOYABLE: indicates story value and interestingness.
    \item Which story uses more FLUENT language: indicates story readability and grammaticality.
\end{itemize}
Similar questions have been used in evaluations of other story generation systems \cite[cf.][]{tambwekar2018controllable,ammanabrolu2020story,ammanabrolu2020automated,Castricato2021Five}. Each pairwise comparison is seen by at least 5 participants.

We conduct our studies using the Cloud Research crowdsourcing platform to interface with Amazon Mechanical Turk \cite{litman2017turkprime}. Only those who pass a screening question are qualified for the study. Participants must also explain their preferences for each comparison with more than $50$ characters of free text. 
We manually verify screening question responses to qualify participants and disregard data for those who fail the screening. 
All crowdsourcing studies we conducted were approved by our institution's Institutional Review Board (IRB).
We recruited 86 participants from the United States, paying \$11.7 per hour on average. Only those with HIT approval rate above 90\% and have over 1000 HITs approved were selected. Average inter-annotator agreement, measured by Fleiss' kappa \cite{fleiss1971measuring}, is > 0.2 (fair); a more detailed breakdown by experiment can be found in \autoref{tab:CAST_results_majority}. 
Further details are provided in Appendix~\ref{app:eva}.
 
We randomly select a subset of first-sentences from the test sets of each dataset---$20$ each of 1-character and 2-character prompts from ROCStories, $10$ prompts from WP, and $10$ from FT. We use these sentences to generate a story continuation of 4 sentences from each system.\footnote{We seed prompts and goal sentences for C2PO. For 2-character stories, we use the interleaving story generation method described in \sect{sec:lm}.}
We recruited $86$ participants on a crowdsourcing platform.
Each participant answered the four pairwise comparison questions (\sect{sec:metrics}) on a randomly selected subset of $5$ story pairs, comprised of one story from CAST and one from one of the baselines.

\subsection{Results}
\label{sec:cast_exp}
The results are shown in \autoref{tab:CAST_results} (top) where we detail the percentage of times human participants choose the story from one system over another for each dimension in the questionnaire.
We indicate when results are significant at $p<0.05$ and $p<0.01$ confidence levels.
Generally, participants strongly preferred stories generated by CAST to those generated by alternatives. 

Compared with \citeauthor{guan2020knowledge}~\shortcite{guan2020knowledge}, CAST is able to find a commonsense inference link to develop the stories on ROCStories prompts, which makes it much more coherent and stay on one single topic.
Human participants state in their response that stories generated by CAST have better commonsense flow and make more sense.
Stories generated by CAST is also more enjoyable and fluent because of its high coherence.\footnote{We observe that these four dimensions are highly, positively correlated using Spearman’s Rank Order Correlation (See Appendix~\ref{app:correlation}).} 

Since COMET is trained on ROCStories, we also seek whether CAST works on other datasets.
We compared CAST to \citeauthor{goldfarb2020content}~\shortcite{goldfarb2020content} on Writing Prompts, which contain longer and more complicated stories.
CAST with language model fine-tuned on Writing Prompts outperforms \citeauthor{goldfarb2020content}~\shortcite{goldfarb2020content} in ``Logical Sense'', ``Single Topic'' and ``Enjoyable'' dimensions.
On the topic of fluency, CAST is preferred but the result is not statistically significant when ties are considered.
Human participants stated that they found stories generated by CAST is much easier to follow and they are built on a single topic.
Because \citeauthor{goldfarb2020content}~\shortcite{goldfarb2020content} applies BART~\cite{lewis2020bart} to generate plots, which cannot ensure commonsense like CAST.

C2PO is also built on COMET to conduct a  bi-directional search from a given start event and a given end event, which makes it as a strong baseline to compare.
We follow \citet{ammanabrolu2020automated} to extract high-level plots from fairy tale stories as prompts and goals\footnote{We only provide prompts to CAST.} for evaluation.
CAST outperforms C2PO on all dimensions, because we apply a harder commonsense constraints on continuation generation than C2PO, which produce a more coherent and on-topic story.
We anecdotally observe that CAST generates more diverse stories than C2PO because of templated and limited range of COMET, which we only use for filtering whereas C2PO uses it for sentence generation.

We conclude that CAST is able to produce a much more coherent, on-topic, enjoyable and fluent story than strong baselines.
It also has the advantage over  \citeauthor{goldfarb2020content}~\shortcite{goldfarb2020content} and C2PO for choosing characters in the story continuations, which makes CAST able to produce single- or two-character stories.

\section{Conclusions}

Neural language models generate content based on the likelihood of tokens given a historical context.
Human readers, on the other hand, use complex inferential processes to connect the relationships between events.
This mismatch between generative models and reader comprehension is one of the reasons why stories generated by neural language models lose coherence over time.

Our CAST system is a straightforward approach to enforce the constraint that a language model only generate continuations that cognitive psychology tells us will be more comprehensible.
The CAST method provides hard constraints to neural language model generation that results in greater story coherence, a result that holds in multiple storytelling domains. We find that perceived story enjoyability and fluency are tied to making logical sense, tracking character goals, and staying on topic; our system excels in all four of these areas.

\section{Acknowledgements}
This work was done while SW was at the Georgia Institute of Technology.

\section{Limitations}
The primary data source of our paper is the ROCStories dataset. ROCStories consists of many event-centric narratives which, while often used in story generation research, is still not representative of complex, realistic narratives. 
This may give $\text{COMET}^{20}_{20}$~\cite{Hwang2021COMETATOMIC2O} an advantage in making inferences that are used for filtering.

$\text{COMET}^{20}_{20}$ requires a clearly identifiable actor in each sentence in order to make commonsense inferences for that actors.
Thus our language model---by virtue of fine-tuning on ROCStories---produces sentences (events) that have an identifiable character peforming an action.
Stories can have more complex expository text.
Narratologists---those that study narratives---often distinguish between {\em events}---text that implies a change to the world and thus drive the story forward---and {\em exposition}---text that describes elements of the story world without changing it.

The performance of CAST is tied to the inference abilities of $\text{COMET}^{20}_{20}$. 
As such, the types of errors that COMET is prone to are also the types of errors that our system is prone to.
We invite readers to review the discussion in \citet{Hwang2021COMETATOMIC2O} for more detailed analysis of commonsense inference errors.
As more advanced commonsense inference models develop, CAST-like approaches will benefit from the improved state-of-the-art.
CAST can easily switch to new generative language models or new commonsense inference model.

Restricted by the filter---$\text{COMET}^{20}_{20}$---our system works mostly for narratives with event-centric
commonsense knowledge.
Even though we processed the datasets (Appendix~\ref{app:data}) to decrease the gender biases, there is no guarantee to entirely eliminate these biases.

CAST produces stories by chaining sentence-level COMET inferences to track important implicit elements of the story between adjacent sentences.
We make a Markovian assumption by only comparing the currently generated event to the most recent event.
Stories are arguably non-Markovian and can have complex, interleaving chains of inference;
despite the assumption, we find in practice that it
enforces global coherence quite successfully (see Single Topic metric in Table 4).
Future work may relax this constraint by keeping track of wants/needs/etc from previous sentences against which to match. 
One would need to solve the problem of deciding when wants/needs/etc should expire because they are no longer applicable.

\bibliography{anthology}

\begin{thebibliography}{49}
\expandafter\ifx\csname natexlab\endcsname\relax\def\natexlab#1{#1}\fi

\bibitem[{Ammanabrolu et~al.(2021)Ammanabrolu, Cheung, Broniec, and Riedl}]{ammanabrolu2020automated}
Prithviraj Ammanabrolu, Wesley Cheung, William Broniec, and Mark~O Riedl. 2021.
\newblock \href {https://arxiv.org/abs/2009.00829} {Automated storytelling via causal, commonsense plot ordering}.
\newblock In \emph{Proceedings of the AAAI Conference on Artificial Intelligence}, volume~35, pages 5859--5867.

\bibitem[{Ammanabrolu et~al.(2020{\natexlab{a}})Ammanabrolu, Cheung, Tu, Broniec, and Riedl}]{ammanabrolu2020bringing}
Prithviraj Ammanabrolu, Wesley Cheung, Dan Tu, William Broniec, and Mark Riedl. 2020{\natexlab{a}}.
\newblock \href {https://dl.acm.org/doi/pdf/10.5555/3505464.3505465} {Bringing stories alive: Generating interactive fiction worlds}.
\newblock In \emph{Proceedings of the AAAI Conference on Artificial Intelligence and Interactive Digital Entertainment (AIIDE)}, volume~16, pages 3--9.

\bibitem[{Ammanabrolu et~al.(2020{\natexlab{b}})Ammanabrolu, Tien, Cheung, Luo, Ma, Martin, and Riedl}]{ammanabrolu2020story}
Prithviraj Ammanabrolu, Ethan Tien, Wesley Cheung, Zhaochen Luo, William Ma, Lara~J Martin, and Mark~O Riedl. 2020{\natexlab{b}}.
\newblock \href {https://arxiv.org/pdf/1909.03480.pdf} {Story realization: Expanding plot events into sentences}.
\newblock In \emph{Proceedings of the AAAI Conference on Artificial Intelligence}, volume~34, pages 7375--7382.

\bibitem[{Bosselut et~al.(2019)Bosselut, Rashkin, Sap, Malaviya, Celikyilmaz, and Choi}]{bosselut2019comet}
Antoine Bosselut, Hannah Rashkin, Maarten Sap, Chaitanya Malaviya, Asli Celikyilmaz, and Yejin Choi. 2019.
\newblock \href {https://doi.org/10.18653/v1/P19-1470} {{COMET}: Commonsense transformers for automatic knowledge graph construction}.
\newblock In \emph{Proceedings of the 57th Annual Meeting of the Association for Computational Linguistics}, pages 4762--4779, Florence, Italy. Association for Computational Linguistics.

\bibitem[{Brahman and Chaturvedi(2020)}]{brahman-chaturvedi-2020-modeling}
Faeze Brahman and Snigdha Chaturvedi. 2020.
\newblock \href {https://doi.org/10.18653/v1/2020.emnlp-main.426} {Modeling protagonist emotions for emotion-aware storytelling}.
\newblock In \emph{Proceedings of the 2020 Conference on Empirical Methods in Natural Language Processing (EMNLP)}, pages 5277--5294, Online. Association for Computational Linguistics.

\bibitem[{Brown et~al.(2020)Brown, Mann, Ryder, Subbiah, Kaplan, Dhariwal, Neelakantan, Shyam, Sastry, Askell, Agarwal, Herbert-Voss, Krueger, Henighan, Child, Ramesh, Ziegler, Wu, Winter, Hesse, Chen, Sigler, Litwin, Gray, Chess, Clark, Berner, McCandlish, Radford, Sutskever, and Amodei}]{brown2020language}
Tom~B. Brown, Benjamin Mann, Nick Ryder, Melanie Subbiah, Jared Kaplan, Prafulla Dhariwal, Arvind Neelakantan, Pranav Shyam, Girish Sastry, Amanda Askell, Sandhini Agarwal, Ariel Herbert-Voss, Gretchen Krueger, T.~J. Henighan, Rewon Child, Aditya Ramesh, Daniel~M. Ziegler, Jeff Wu, Clemens Winter, Christopher Hesse, Mark Chen, Eric Sigler, Mateusz Litwin, Scott Gray, Benjamin Chess, Jack Clark, Christopher Berner, Sam McCandlish, Alec Radford, Ilya Sutskever, and Dario Amodei. 2020.
\newblock \href {https://proceedings.neurips.cc/paper/2020/file/1457c0d6bfcb4967418bfb8ac142f64a-Paper.pdf} {Language models are few-shot learners}.
\newblock \emph{Advances in Neural Information Processing Systems}, 33:1877--1901.

\bibitem[{Caglayan et~al.(2020)Caglayan, Madhyastha, and Specia}]{caglayan-etal-2020-curious}
Ozan Caglayan, Pranava Madhyastha, and Lucia Specia. 2020.
\newblock \href {https://doi.org/10.18653/v1/2020.coling-main.210} {Curious case of language generation evaluation metrics: A cautionary tale}.
\newblock In \emph{Proceedings of the 28th International Conference on Computational Linguistics}, pages 2322--2328, Barcelona, Spain (Online). International Committee on Computational Linguistics.

\bibitem[{Castricato et~al.(2021)Castricato, Frazier, Balloch, and Riedl}]{Castricato2021Five}
Louis Castricato, Spencer Frazier, Jonathan Balloch, and Mark~O. Riedl. 2021.
\newblock \href {https://par.nsf.gov/servlets/purl/10249509} {Tell me a story like i'm five: Story generation via question answering}.
\newblock In \emph{Proceedings of the 3rd Workshop on Narrative Understanding}, Virtual. Association for Computational Linguistics.

\bibitem[{Celikyilmaz et~al.(2020)Celikyilmaz, Clark, and Gao}]{celikyilmaz2020evaluation}
Asli Celikyilmaz, Elizabeth Clark, and Jianfeng Gao. 2020.
\newblock \href {https://arxiv.org/pdf/2006.14799.pdf} {Evaluation of text generation: A survey}.
\newblock \emph{arXiv preprint arXiv:2006.14799}.

\bibitem[{Clark et~al.(2018)Clark, Ji, and Smith}]{clark2018}
Elizabeth Clark, Yangfeng Ji, and Noah~A. Smith. 2018.
\newblock \href {https://doi.org/10.18653/v1/N18-1204} {Neural text generation in stories using entity representations as context}.
\newblock In \emph{Proceedings of the 2018 Conference of the North {A}merican Chapter of the Association for Computational Linguistics: Human Language Technologies, Volume 1 (Long Papers)}, pages 2250--2260, New Orleans, Louisiana. Association for Computational Linguistics.

\bibitem[{De~Smedt and Daelemans(2012)}]{de2012pattern}
Tom De~Smedt and Walter Daelemans. 2012.
\newblock \href {https://www.jmlr.org/papers/volume13/desmedt12a/desmedt12a.pdf} {Pattern for python}.
\newblock \emph{The Journal of Machine Learning Research}, 13(1):2063--2067.

\bibitem[{Fan et~al.(2018)Fan, Lewis, and Dauphin}]{fan2018hierarchical}
Angela Fan, Mike Lewis, and Yann Dauphin. 2018.
\newblock \href {https://doi.org/10.18653/v1/P18-1082} {Hierarchical neural story generation}.
\newblock In \emph{Proceedings of the 56th Annual Meeting of the Association for Computational Linguistics (Volume 1: Long Papers)}, pages 889--898, Melbourne, Australia. Association for Computational Linguistics.

\bibitem[{Fan et~al.(2019)Fan, Lewis, and Dauphin}]{fan2019strategies}
Angela Fan, Mike Lewis, and Yann Dauphin. 2019.
\newblock \href {https://doi.org/10.18653/v1/P19-1254} {Strategies for structuring story generation}.
\newblock In \emph{Proceedings of the 57th Annual Meeting of the Association for Computational Linguistics}, pages 2650--2660, Florence, Italy. Association for Computational Linguistics.

\bibitem[{Fleiss(1971)}]{fleiss1971measuring}
Joseph~L Fleiss. 1971.
\newblock \href {http://www.wpic.pitt.edu/research/biometrics/Publications/Biometrics%20Archives%20PDF/395-1971%20Fleiss0001.pdf} {Measuring nominal scale agreement among many raters.}
\newblock \emph{Psychological bulletin}, 76(5):378.

\bibitem[{Gerv{\'a}s(2013)}]{gervas2013propp}
Pablo Gerv{\'a}s. 2013.
\newblock \href {https://drops.dagstuhl.de/opus/volltexte/2013/4156/pdf/p106-gervas.pdf} {Propp's morphology of the folk tale as a grammar for generation}.
\newblock In \emph{2013 Workshop on Computational Models of Narrative}.

\bibitem[{Gerv{\'a}s(2014)}]{gervas2014composing}
Pablo Gerv{\'a}s. 2014.
\newblock \href {https://academic.oup.com/dsh/article/29/4/511/982149} {Composing narrative discourse for stories of many characters: a case study over a chess game}.
\newblock \emph{Literary and Linguistic Computing}, 29(4):511--531.

\bibitem[{Goldfarb-Tarrant et~al.(2020)Goldfarb-Tarrant, Chakrabarty, Weischedel, and Peng}]{goldfarb2020content}
Seraphina Goldfarb-Tarrant, Tuhin Chakrabarty, Ralph Weischedel, and Nanyun Peng. 2020.
\newblock \href {https://doi.org/10.18653/v1/2020.emnlp-main.351} {Content planning for neural story generation with aristotelian rescoring}.
\newblock In \emph{Proceedings of the 2020 Conference on Empirical Methods in Natural Language Processing (EMNLP)}, pages 4319--4338, Online. Association for Computational Linguistics.

\bibitem[{Graesser et~al.(1991)Graesser, Lang, and Roberts}]{graesser91}
Art Graesser, Kathy~L. Lang, and Richard~M. Roberts. 1991.
\newblock \href {https://web.p.ebscohost.com/ehost/pdfviewer/pdfviewer?vid=0&sid=00cdc7c3-f9fb-4489-af38-6801c398c0db\%40redis} {Question answering in the context of stories}.
\newblock \emph{Journal of Experimental Psychology: General}, 120(3):254--277.

\bibitem[{Graesser et~al.(1994)Graesser, Singer, and Trabasso}]{graesser94}
Art Graesser, Murray Singer, and Tom Trabasso. 1994.
\newblock \href {https://web.archive.org/web/20170809034152id_/https://www.macalester.edu/academics/psychology/rali-lab/articles/Graesser,\%20Singer,\%20&\%20Trabasso\%20(1994)\%20Constructing\%20inferences\%20during\%20narrative\%20text\%20comprehension.pdf} {Constructing inferences during narrative text comprehension}.
\newblock \emph{Psychological Review}, 101(3):371--395.

\bibitem[{Guan et~al.(2020)Guan, Huang, Zhao, Zhu, and Huang}]{guan2020knowledge}
Jian Guan, Fei Huang, Zhihao Zhao, Xiaoyan Zhu, and Minlie Huang. 2020.
\newblock \href {https://doi.org/10.1162/tacl_a_00302} {{A Knowledge-Enhanced Pretraining Model for Commonsense Story Generation}}.
\newblock \emph{Transactions of the Association for Computational Linguistics}, 8:93--108.

\bibitem[{Guan et~al.(2019)Guan, Wang, and Huang}]{guan2019story}
Jian Guan, Yansen Wang, and Minlie Huang. 2019.
\newblock \href {https://dl.acm.org/doi/pdf/10.1609/aaai.v33i01.33016473} {Story ending generation with incremental encoding and commonsense knowledge}.
\newblock In \emph{Proceedings of the AAAI Conference on Artificial Intelligence}, volume~33, pages 6473--6480.

\bibitem[{Holtzman et~al.(2019)Holtzman, Buys, Du, Forbes, and Choi}]{holtzman2019curious}
Ari Holtzman, Jan Buys, Li~Du, Maxwell Forbes, and Yejin Choi. 2019.
\newblock \href {https://openreview.net/pdf?id=rygGQyrFvH} {The curious case of neural text degeneration}.
\newblock In \emph{Proceedings of the International Conference on Learning Representations (ICLR)}.

\bibitem[{Hwang et~al.(2021)Hwang, Bhagavatula, {Le Bras}, Da, Sakaguchi, Bosselut, and Choi}]{Hwang2021COMETATOMIC2O}
Jena~D. Hwang, Chandra Bhagavatula, Ronan {Le Bras}, Jeff Da, Keisuke Sakaguchi, Antoine Bosselut, and Yejin Choi. 2021.
\newblock \href {https://arxiv.org/pdf/2010.05953.pdf} {{COMET}-{ATOMIC} 2020: On symbolic and neural commonsense knowledge graphs}.
\newblock In \emph{Proceedings of the AAAI Conference on Artificial Intelligence}, volume~35, pages 6384--6392.

\bibitem[{Khalifa et~al.(2017)Khalifa, Barros, and Togelius}]{khalifa2017deeptingle}
Ahmed Khalifa, Gabriella~AB Barros, and Julian Togelius. 2017.
\newblock \href {https://arxiv.org/pdf/1705.03557.pdf} {Deep{T}ingle}.
\newblock In \emph{Proceedings of the 8th International Conference on Computational Creativity}.

\bibitem[{Lewis et~al.(2020)Lewis, Liu, Goyal, Ghazvininejad, Mohamed, Levy, Stoyanov, and Zettlemoyer}]{lewis2020bart}
Mike Lewis, Yinhan Liu, Naman Goyal, Marjan Ghazvininejad, Abdelrahman Mohamed, Omer Levy, Veselin Stoyanov, and Luke Zettlemoyer. 2020.
\newblock \href {https://doi.org/10.18653/v1/2020.acl-main.703} {{BART}: Denoising sequence-to-sequence pre-training for natural language generation, translation, and comprehension}.
\newblock In \emph{Proceedings of the 58th Annual Meeting of the Association for Computational Linguistics}, pages 7871--7880, Online. Association for Computational Linguistics.

\bibitem[{Lin et~al.(2022)Lin, Cao, Huang, Li, Hu, Wen, and Wang}]{lin2022inferring}
Li~Lin, Yixin Cao, Lifu Huang, Shuang Li, Xuming Hu, Lijie Wen, and Jianmin Wang. 2022.
\newblock Inferring commonsense explanations as prompts for future event generation.
\newblock \emph{arXiv preprint arXiv:2201.07099}.

\bibitem[{Litman et~al.(2017)Litman, Robinson, and Abberbock}]{litman2017turkprime}
Leib Litman, Jonathan Robinson, and Tzvi Abberbock. 2017.
\newblock \href {https://link.springer.com/content/pdf/10.3758/s13428-016-0727-z.pdf} {Turkprime. com: A versatile crowdsourcing data acquisition platform for the behavioral sciences}.
\newblock \emph{Behavior research methods}, 49(2):433--442.

\bibitem[{Martin et~al.(2018)Martin, Ammanabrolu, Wang, Hancock, Singh, Harrison, and Riedl}]{martin2018event}
Lara Martin, Prithviraj Ammanabrolu, Xinyu Wang, William Hancock, Shruti Singh, Brent Harrison, and Mark Riedl. 2018.
\newblock \href {https://arxiv.org/abs/1706.01331} {Event representations for automated story generation with deep neural nets}.
\newblock In \emph{Proceedings of the AAAI Conference on Artificial Intelligence}, volume~32, pages 868--875.

\bibitem[{Miller(1995)}]{miller1995wordnet}
George~A Miller. 1995.
\newblock \href {https://dl.acm.org/doi/pdf/10.1145/219717.219748?casa_token=JTDujopxvZwAAAAA:ttOmsIcHCEQ3NEh6JSlcSm82b7IbGJlOaRz8xz1UrlQfKyOAMO3M-hxUQ5B-P_Noze62v9DtTl4f7w} {{W}ord{N}et: a lexical database for english}.
\newblock \emph{Communications of the ACM}, 38(11):39--41.

\bibitem[{Mostafazadeh et~al.(2016)Mostafazadeh, Chambers, He, Parikh, Batra, Vanderwende, Kohli, and Allen}]{mostafazadeh2016corpus}
Nasrin Mostafazadeh, Nathanael Chambers, Xiaodong He, Devi Parikh, Dhruv Batra, Lucy Vanderwende, Pushmeet Kohli, and James Allen. 2016.
\newblock \href {https://doi.org/10.18653/v1/N16-1098} {A corpus and cloze evaluation for deeper understanding of commonsense stories}.
\newblock In \emph{Proceedings of the 2016 Conference of the North {A}merican Chapter of the Association for Computational Linguistics: Human Language Technologies}, pages 839--849, San Diego, California. Association for Computational Linguistics.

\bibitem[{Ouyang and McKeown(2015)}]{ouyang-mckeown-2015-modeling}
Jessica Ouyang and Kathleen McKeown. 2015.
\newblock \href {https://doi.org/10.18653/v1/D15-1257} {Modeling reportable events as turning points in narrative}.
\newblock In \emph{Proceedings of the 2015 Conference on Empirical Methods in Natural Language Processing}, pages 2149--2158, Lisbon, Portugal. Association for Computational Linguistics.

\bibitem[{Papineni et~al.(2002)Papineni, Roukos, Ward, and Zhu}]{papineni2002bleu}
Kishore Papineni, Salim Roukos, Todd Ward, and Wei-Jing Zhu. 2002.
\newblock \href {https://doi.org/10.3115/1073083.1073135} {{B}leu: a method for automatic evaluation of machine translation}.
\newblock In \emph{Proceedings of the 40th Annual Meeting of the Association for Computational Linguistics}, pages 311--318, Philadelphia, Pennsylvania, USA. Association for Computational Linguistics.

\bibitem[{Paul and Frank(2021)}]{paul-frank-2021-coins}
Debjit Paul and Anette Frank. 2021.
\newblock \href {https://doi.org/10.18653/v1/2021.acl-long.395} {{COINS}: Dynamically generating {CO}ntextualized inference rules for narrative story completion}.
\newblock In \emph{Proceedings of the 59th Annual Meeting of the Association for Computational Linguistics and the 11th International Joint Conference on Natural Language Processing (Volume 1: Long Papers)}, pages 5086--5099, Online. Association for Computational Linguistics.

\bibitem[{Peng et~al.(2018)Peng, Ghazvininejad, May, and Knight}]{peng-etal-2018-towards}
Nanyun Peng, Marjan Ghazvininejad, Jonathan May, and Kevin Knight. 2018.
\newblock \href {https://doi.org/10.18653/v1/W18-1505} {Towards controllable story generation}.
\newblock In \emph{Proceedings of the First Workshop on Storytelling}, pages 43--49, New Orleans, Louisiana. Association for Computational Linguistics.

\bibitem[{Peng and Sollami(2022)}]{peng2022xfboost}
Xiangyu Peng and Michael Sollami. 2022.
\newblock \href {https://arxiv.org/pdf/2202.08124.pdf} {{XFB}oost: Improving text generation with controllable decoders}.
\newblock \emph{arXiv preprint arXiv:2202.08124}.

\bibitem[{Purdy et~al.(2018)Purdy, Wang, He, and Riedl}]{purdy2018predicting}
Christopher Purdy, Xinyu Wang, Larry He, and Mark Riedl. 2018.
\newblock \href {https://www.aaai.org/ocs/index.php/AIIDE/AIIDE18/paper/viewFile/18106/17228} {Predicting generated story quality with quantitative measures}.
\newblock In \emph{Proceedings of the AAAI Conference on Artificial Intelligence and Interactive Digital Entertainment (AIIDE)}, volume~14, pages 95--101.

\bibitem[{Radford et~al.(2019)Radford, Wu, Child, Luan, Amodei, and Sutskever}]{radford2019language}
Alec Radford, Jeffrey Wu, Rewon Child, David Luan, Dario Amodei, and Ilya Sutskever. 2019.
\newblock \href {https://cdn.openai.com/better-language-models/language_models_are_unsupervised_multitask_learners.pdf} {Language models are unsupervised multitask learners}.

\bibitem[{Rashkin et~al.(2018)Rashkin, Bosselut, Sap, Knight, and Choi}]{rashkin-etal-2018-modeling}
Hannah Rashkin, Antoine Bosselut, Maarten Sap, Kevin Knight, and Yejin Choi. 2018.
\newblock \href {https://doi.org/10.18653/v1/P18-1213} {Modeling naive psychology of characters in simple commonsense stories}.
\newblock In \emph{Proceedings of the 56th Annual Meeting of the Association for Computational Linguistics (Volume 1: Long Papers)}, pages 2289--2299, Melbourne, Australia. Association for Computational Linguistics.

\bibitem[{Rashkin et~al.(2020)Rashkin, Celikyilmaz, Choi, and Gao}]{rashkin2020plotmachines}
Hannah Rashkin, Asli Celikyilmaz, Yejin Choi, and Jianfeng Gao. 2020.
\newblock \href {https://doi.org/10.18653/v1/2020.emnlp-main.349} {{P}lot{M}achines: Outline-conditioned generation with dynamic plot state tracking}.
\newblock In \emph{Proceedings of the 2020 Conference on Empirical Methods in Natural Language Processing (EMNLP)}, pages 4274--4295, Online. Association for Computational Linguistics.

\bibitem[{Reimers and Gurevych(2019)}]{reimers-2019-sentence-bert}
Nils Reimers and Iryna Gurevych. 2019.
\newblock \href {https://doi.org/10.18653/v1/D19-1410} {Sentence-{BERT}: Sentence embeddings using {S}iamese {BERT}-networks}.
\newblock In \emph{Proceedings of the 2019 Conference on Empirical Methods in Natural Language Processing and the 9th International Joint Conference on Natural Language Processing (EMNLP-IJCNLP)}, pages 3982--3992, Hong Kong, China. Association for Computational Linguistics.

\bibitem[{Riedl(2016)}]{riedl:chi-hcml2016}
Mark~O. Riedl. 2016.
\newblock \href {http://www.cc.gatech.edu/~riedl/pubs/chi-hcml16.pdf} {Computational narrative intelligence: A human-centered goal for artificial intelligence}.
\newblock In \emph{Proceedings of the CHI 2016 Workshop on Human Centered Machine Learning}.

\bibitem[{Riedl and Young(2010)}]{riedl:jair2010}
Mark~O. Riedl and R.~Michael Young. 2010.
\newblock \href {https://arxiv.org/abs/1401.3841} {Narrative planning: Balancing plot and character}.
\newblock \emph{Journal of Artificial Intelligence Research}, 39:217--268.

\bibitem[{Roemmele(2016)}]{roemmele2016writing}
Melissa Roemmele. 2016.
\newblock \href {https://www.aaai.org/ocs/index.php/AAAI/AAAI16/paper/viewFile/11966/12271} {Writing stories with help from recurrent neural networks}.
\newblock In \emph{Proceedings of the AAAI Conference on Artificial Intelligence}, volume~30, pages 4311--4312.

\bibitem[{Sap et~al.(2019)Sap, Le~Bras, Allaway, Bhagavatula, Lourie, Rashkin, Roof, Smith, and Choi}]{sap2018atomic}
Maarten Sap, Ronan Le~Bras, Emily Allaway, Chandra Bhagavatula, Nicholas Lourie, Hannah Rashkin, Brendan Roof, Noah~A Smith, and Yejin Choi. 2019.
\newblock \href {https://arxiv.org/pdf/1811.00146.pdf} {{ATOMIC}: An atlas of machine commonsense for if-then reasoning}.
\newblock In \emph{Proceedings of the AAAI Conference on Artificial Intelligence}, volume~33, pages 3027--3035.

\bibitem[{Tambwekar et~al.(2019)Tambwekar, Dhuliawala, Martin, Mehta, Harrison, and Riedl}]{tambwekar2018controllable}
Pradyumna Tambwekar, Murtaza Dhuliawala, Lara~J Martin, Animesh Mehta, Brent Harrison, and Mark~O Riedl. 2019.
\newblock \href {https://www.ijcai.org/proceedings/2019/0829.pdf} {Controllable neural story plot generation via reward shaping}.
\newblock In \emph{Proceedings of the International Joint Conference on Artificial Intelligence (IJCAI)}, volume~28, pages 5982--5988.

\bibitem[{Trabasso and Van Den~Broek(1985)}]{trabasso1985causal}
Tom Trabasso and Paul Van Den~Broek. 1985.
\newblock \href {https://www.researchgate.net/profile/Paul-Van-Den-Broek/publication/222232677_Causal_Thinking_and_the_Representation_of_Narrative_Events/links/59f6f2b2a6fdcc075ec61c75/Causal-Thinking-and-the-Representation-of-Narrative-Events.pdf} {Causal thinking and the representation of narrative events}.
\newblock \emph{Journal of memory and language}, 24(5):612--630.

\bibitem[{van~der Lee et~al.(2021)van~der Lee, Gatt, van Miltenburg, and Krahmer}]{van2021human}
Chris van~der Lee, Albert Gatt, Emiel van Miltenburg, and Emiel Krahmer. 2021.
\newblock \href {https://www.sciencedirect.com/sdfe/reader/pii/S088523082030084X/pdf} {Human evaluation of automatically generated text: Current trends and best practice guidelines}.
\newblock \emph{Computer Speech \& Language}, 67:101151.

\bibitem[{Yao et~al.(2019)Yao, Peng, Weischedel, Knight, Zhao, and Yan}]{yao2019plan}
Lili Yao, Nanyun Peng, Ralph Weischedel, Kevin Knight, Dongyan Zhao, and Rui Yan. 2019.
\newblock \href {https://arxiv.org/pdf/1811.05701v3.pdf} {{P}lan-and-{W}rite: Towards better automatic storytelling}.
\newblock In \emph{Proceedings of the AAAI Conference on Artificial Intelligence}, volume~33, pages 7378--7385.

\bibitem[{Zhu et~al.(2018)Zhu, Lu, Zheng, Guo, Zhang, Wang, and Yu}]{zhu2018texygen}
Yaoming Zhu, Sidi Lu, Lei Zheng, Jiaxian Guo, Weinan Zhang, Jun Wang, and Yong Yu. 2018.
\newblock \href {https://dl.acm.org/doi/pdf/10.1145/3209978.3210080?casa_token=csRcImwMJ1YAAAAA:EpFiss9b0B7vezWl3tst1cSG-1U-ieFH01nOd9Y8sbkdEIdq_4-I4Em6dcX_-HijMjX9oCOj_FhqQg} {Texygen: A benchmarking platform for text generation models}.
\newblock In \emph{Proceedings of the International ACM SIGIR Conference on Research \& Development in Information Retrieval}, volume~41, pages 1097--1100.

\end{thebibliography}
\bibliographystyle{acl_natbib}

\appendix
\section{Implementation Details} \label{sec:appendix}

\subsection{Data} 
\label{app:data}
We use the preprocessed ROCStories corpus of 5-sentence stories (588,966 stories) and joint ATOMIC and ConceptNet dataset converted to template sentences (1,174,267 train/66,856 dev/73,083 test), both provided by \citeauthor{guan2020knowledge}\footnote{\url{https://cloud.tsinghua.edu.cn/d/670f7787b6554f308226/}} 
We shuffle and split the ROCStories dataset into 80\% (78,528) train and 20\% (19,633) test sets.

Following \citeauthor{guan2020knowledge}~\shortcite{guan2020knowledge}, character names in the ROCStories corpus are replaced with \texttt{[MALE]} or \texttt{[FEMALE]} tags.
In order to remove gender bias, we replace \texttt{[MALE]}, \texttt{[FEMALE]}, or \texttt{[NEUTRAL]} tags with \texttt{[Char\_1]}, \texttt{[Char\_2]}, or \texttt{[Char\_3]} tags.
This prevents skewed predictions due to the presence of certain names in a small dataset such as ROCStories and also allows us to focus on 2-character stories without having to perform NER on generated sentences to remove extraneously generated names outside of the two main characters.
It also allows a direct comparison to prior work. 
After a story is generated, we replace the character tags with user-inputted names, assuming the subject and object of the first sentence are the subsequently-generated tags.

\subsection{Models}
\label{app:baselines}

Following \citeauthor{guan2020knowledge}, we use the small version of GPT-2 with 124M parameters as the base for our fine-tuned models. 
When fine-tuning GPT-2 on either ROCStories and the commonsense knowledge resources (done separately), we train with a learning rate of 0.00005, and using the Adam optimizer with gradient clipping at a max norm of 1. 
CAST and \citet{guan2020knowledge} were trained on single GeForce RTX 2080 GPUs in Pytorch using the Huggingface Transformers library.\footnote{\url{https://huggingface.co/transformers/}} We replicate the multi-task baseline of \citeauthor{guan2020knowledge} in Tensorflow using their provided code.\footnote{\url{https://github.com/thu-coai/CommonsenseStoryGen}} 
We train with early stopping on the dev set (80\% train,10\% dev,10\% test split) loss with a patience of 10 epochs. Both models converge within 1-2 epochs. All other training details are kept the same.
We use top-p sampling \cite{holtzman2019curious} with a value of $0.9$, a temperature of $1$, and a max length of 20 tokens per sentence to sample from CAST and \citet{guan2020knowledge}.

For \citet{goldfarb2020content}, we use BART model using the code and parameters published on the paper's public repository \footnote{\href{https://github.com/PlusLabNLP/story-gen-BART}{https://github.com/PlusLabNLP/story-gen-BART}}.

We replicate the C2PO model by \citet{ammanabrolu2020automated} using the code published on the paper's public repository \footnote{\href{https://github.com/rajammanabrolu/C2PO}{https://github.com/rajammanabrolu/C2PO}}.
All the encoder and model checkpoints are provided by the author.

\subsection{Character Conditioned Generation}
\label{app:appendix-prefix}
To enforce the telling of a two-character narrative in an interleaving fashion wherein characters take turns being the subject of each sentence. 
We fine-tune the language model by formulating the input as $*$T$*$ $[s_1,\dots, s_{i-1}]$, where T is the tag denoting the character who is to take a turn, which is determined by Part-Of-Speech Tagger\footnote{\url{https://nlp.stanford.edu/software/tagger.shtml}}.
For example, for the story, ``\texttt{[Char\_1]} was upset with \texttt{[Char\_2]}. Because of this,  \texttt{[Char\_2]} apologized.'', the prompt is formulated as ``* \texttt{[Char\_2]} * \texttt{[Char\_1]} was upset with \texttt{[Char\_2]}.'' The language model is fine-tuned by back-propagating the loss calculated on the sentence ``Because of this,  \texttt{[Char\_2]} apologized.'' At test-time, we generate second sentence candidates until one contains a reference to \texttt{[Char\_2]}.

\subsection{Commonsense Matching Criteria}
\label{app:pairs}
We randomly selected 500 stories from ROCStories\cite{mostafazadeh2016corpus}.
We then use COMET to produce commonsense inference sets for these stories of all the $34$ relations in ATOMIC with beam size of $10$.
Hence, for all the $500 \times 5$ sentences, we obtain $10$ commonsense inference for each type ($34$ types).
For each sentence, we consider the commonsense relation sets of current sentence and its next sentence as a pair.
So we have $32 \times 31$ pairs for each pair of the adjacent sentences.
Then we adopt Sentence-BERT~\cite{reimers-2019-sentence-bert} to encode all these inference and calculate the max cosine similarity of each commonsense inference pair for each adjacence sentence pair.
Inference pairs with over $80\%$ semantic similarities are used as hard constraints via a form of chaining that allows us to filter a set of potential sentence generations to find one that adequately matches the expected inferences.

\subsection{Ablation Study of Commonsense Inferences Matching}
\label{app:ablation}
\begin{table}[!tbh]
\footnotesize
\centering
\setlength\tabcolsep{3.5pt}
\begin{tabular}{c|c|c|c | c  }
\toprule
\textbf{Semantic}   & \textbf{\# of Sentence}  & \textbf{Success} &\textbf{Self $\downarrow$} &\textbf{Self $\downarrow$}   \\ 
\textbf{Similarity} &\textbf{Candidates $\downarrow$}  & \textbf{Rate $\uparrow$} & \textbf{BLEU-2} & \textbf{BLEU-3}  \\
\midrule
0.8     &  11.25 & 98.75\% & .1718 & .0892\\
0.85   &  17.25 &  97.25\% & .1978 & .1140 \\
0.9   &  18.05 &  91.25\% & .2011 & .1216\\

\bottomrule
\end{tabular}
\caption{Ablation study result for semantic similarity. We run CAST without controlling decoding to generate 20 2-char stories in $5$ seeds. \textit{\# of sentence candidates} denotes the average number of sentences candidates generated before finding a matching inference type pair. \textit{Success rate} is the percentage of finding a match within the 50-candidate limit. A lower Self-BLEU score implies more diversity of the document \cite{zhu2018texygen} (see \S\ref{sec:metrics}). 
}
\label{tab:ablation-result}
\end{table}
\begin{table}[!tbh]
\centering
\footnotesize
\setlength\tabcolsep{0.1pt}
\begin{tabular}{p{0.95\linewidth}}
\toprule
\textbf{Seed Prompt:}  \\
Bob was in love with boyfriend Alice.\\
\midrule
\textbf{Semantic Similarity = 0.8:} 
\\
 Alice was in love with Bob since grade school.\\
 One day, Bob decided she would get married with Alice.\\
 Alice proposed to Bob.\\
 Bob and Alice got married.\\
\midrule
\textbf{Semantic Similarity = 0.9:}\\
 Alice wanted to get \textbf{married}.\\ Bob was asked to \textbf{marry} in the article.\\
 Alice prepared to get \textbf{married}.\\ Bob proposed..\\
 \midrule
 \midrule
 \textbf{Seed Prompt:}  \\
Bob 's cousin Alice wanted the kids to play outside in the summer.\\
\midrule
\textbf{Semantic Similarity = 0.8:} 
\\
 But the kids had a crush on Alice and wanted to play in the yard.\\
 Bob had a good plan.\\
 Alice went to the store to get them new toys.\\
 Bob brought them to the car and went home with the new toys.\\

\midrule
\textbf{Semantic Similarity = 0.9:}\\
So Alice took the kids outside.\\ Bob was surprised that Bob cousin was back.\\
Alice loved playing with the kids.\\ Bob was not the only one to play, and the two played happily.\\

\bottomrule
\end{tabular}
\caption{Story examples generated by CAST with different semantic similarity thresholds. Stories generated at 80\% similarity maintains more diversity.}
\label{tab:story_ablation-sim}
\end{table}
\begin{table}[!tbh]
\footnotesize
\centering
\setlength\tabcolsep{3.5pt}
\begin{tabular}{c|c|c|c | c  }
\toprule
\textbf{\# of}   & \textbf{\# of Sentence}  & \textbf{Success} &\textbf{Self $\downarrow$} &\textbf{Self $\downarrow$} \\ 
\textbf{Matching} &\textbf{Candidates $\downarrow$}  & \textbf{Rate $\uparrow$} & \textbf{BLEU-2} & \textbf{BLEU-3} \\
\midrule
3    & 9.78  & 98.67\% & .1559 & .0860\\
4    & 18.95 & 86.83\% & .1832 &  .1282\\
5    & 56.39 & 40.17\% & .1870 & .1356\\
\bottomrule
\end{tabular}
\caption{Ablation study result for required matching inference type pairs in single-character stories. We run CAST without controlling decoding to generate 30 single-character stories in $5$ seeds. \textit{\# of sentence candidates} denotes the average number of sentences candidates generated before finding a matching inference type pair. \textit{Success rate} is the percentage of finding a match within the 50-candidate limit. The number of sentence candidates could 
Failure to find a match within the candidates limit ($50$) will relax the matching constraints to one pair. Hence, the average number of sentences candidates might be over the candidate limit. 
A lower Self-BLEU score implies more diversity of the document \cite{zhu2018texygen} (see \S\ref{sec:metrics}). 
}
\label{tab:ablation-result_number}
\end{table}
\begin{table}[!tbh]
\centering
\footnotesize
\setlength\tabcolsep{0.1pt}
\begin{tabular}{p{0.95\linewidth}}
\toprule
\textbf{Seed Prompt:}  \\
Bob enjoyed long walks on the beach.\\
\midrule
\textbf{$\#$ of Matching = 3:} 
\\
Bob was always healthy and energetic. \\
Bob enjoyed the sun and the heat. \\
One day, Bob decided to take a walk in the beach.\\ Bob had fun at the beach for the whole day.\\

\midrule
\textbf{$\#$ of Matching = 5:}\\
 One day, Bob decided to go for a long walk on the beach.\\
 Bob loved the sun so much, Bob always happy. \\
 Bob enjoyed the sun when Bob walked on the beach.\\ Bob liked Bob walk on the beach.\\

 \midrule
 \midrule
 \textbf{Seed Prompt:}  \\
Bob had just learned how to ride a bike.\\
\midrule
\textbf{$\#$ of Matching = 3:} 
\\
Bob went to the store to buy a new bike. \\
After buying a new bike, Bob went to ride it.\\
Bob rode it to the park.\\ Bob loved his new bike.\\
\midrule
\textbf{$\#$ of Matching = 5:}\\
 Bob mom took Bob on a bike ride.\\
 Bob went on the bike for hours.\\
 Finally Bob was back on Bob bike.\\
 Bob loved riding it.\\

\bottomrule
\end{tabular}
\caption{Story examples generated by CAST with different semantic similarity thresholds. Stories generated at 80\% similarity maintains more diversity.}
\label{tab:story_ablation-num}
\end{table}
We use $80\%$ semantic similarity as our lower-bound. 
Empirically, we find this value best considers the inferences listed in Section~\ref{sec:comet} as matches, but excludes less-related inferences. 
Table~\ref{tab:story_ablation-sim} shows how the threshold affects success rate---the percentage of queries that find a match within $50$ generated candidates---and the diversity of results as measured by self-BLEU score (described in \S\ref{sec:metrics}). 
Each system was conditioned on the same 20 2-character prompts from ROCStories with 5 different random seeds, requiring two of three inference type pairs to match to qualify as a match. Failure to find a match within the candidates limit ($50$) will relax the matching constraints to two pairs. 
Hence, the average number of sentences candidates might be over the candidate limit.
As observed in \ref{tab:ablation-result} , increasing the semantic similarity threshold decreases the success rate in obtaining a matching candidate within the sentence limit, and it results in more repetitive sentences (see Table~\ref{tab:story_ablation-sim}). 

In order to balance computation time and quality of the match, we only require three of five inference type pairs to match between a seed and a candidate sentence when generating single-character stories. 
When requiring five matches when generating single-char story, CAST only finds a ``qualified'' sentence $40\%$ of the time within 50 attempts (see Table~\ref{tab:ablation-result} (bottom), computed at 0.8 semantic similarity).
In practice (see examples in Table~\ref{tab:story_ablation-num}), we find requiring three pairs results in higher quality sentences than if we only require one or two out of three pairs to match, but is significantly more efficient than four or five out of five.

\subsection{Decoding Process Ablation Study}
\label{app:decode}
\begin{table}[!tbh]
\footnotesize
\centering
\setlength\tabcolsep{3.5pt}
\begin{tabular}{c|c|c|c | c  }
\toprule
\textbf{Decoding}   & \textbf{\# of Sentence}  & \textbf{Success} &\textbf{Self $\downarrow$} &\textbf{Self $\downarrow$} \\ 
\textbf{Matching} &\textbf{Candidates $\downarrow$}  & \textbf{Rate $\uparrow$} & \textbf{BLEU-2} & \textbf{BLEU-3} \\
\midrule
True    & 3.21  & 99.50\% & \textbf{.1709} & .0913\\
False    & 11.25 & 98.75\% & .1718 &  \textbf{.0892}\\
\bottomrule
\end{tabular}
\caption{Ablation study result for required controlling decoding stage. We run CAST with or without controlling decoding to generate 20 multiple-character stories in $5$ seeds. \textit{\# of sentence candidates} denotes the average number of sentences candidates generated before finding a matching inference type pair. \textit{Success rate} is the percentage of finding a match within the 50-candidate limit. A lower Self-BLEU score implies more diversity of the document \cite{zhu2018texygen}. 
}
\label{tab:ablation-result_decode}
\end{table}
In order to increase the probability of finding a match in Section~\ref{sec:comet}, inspired by \citeauthor{peng2022xfboost}~\shortcite{peng2022xfboost}, we use commonsense inferences of prompt sentences as lexical constraints to control the generation decoding process.
We run a ablation test to validate this component of CAST.
Table~\ref{tab:ablation-result_decode} shows that after applying commonsense inferences of prompt sentences as lexical constraints to control the generation decoding process, CAST successfully find a match in the average of 3 candidates. 
At the same time, self-BLEU score did not show any statistical difference. Hence, we adopt decoding technique in CAST. 

\subsection{CAST}
When producing commonsense inferences from COMET, we use ``beam-5" setting to generate 5 inferences for each inference type, which results in a higher percent of matched inferences in our preliminary experiments. We also qualitatively observe that matching on a larger set of inferences (as shown in the demo\footnote{\url{https://mosaickg.apps.allenai.org/comet_atomic}}) more often results in at least one or a few high-quality inferences, due to COMET having some error.

As mentioned in the body of the text, we use a semantic similarity threshold of 80\% and require 3 of 5 inferences to match when generating single-character stories. 
Runtime is feasible due to matching on three out of five inference filters and using the 5-beam COMET output.
However, in some rare cases, no matching next-sentence candidate can be found. 
If no qualified sentence is found after 50 generated candidates, in order to avoid potentially infinite search we loosen the filtering strength to match only one pair of inferences.
We also report the majority vote of experiments in Table~\ref{tab:CAST_results_majority}.

\begin{table*}[tbh!]
\footnotesize
\centering
\setlength\tabcolsep{0.1pt}
\begin{tabular}{c|c|c|lll|lll|lll|lll|c}
    \toprule
    \multirow{2}{*}{\textbf{Models}} & \textbf{Data}&
    \textbf{Num} & \multicolumn{3}{c|}{\textbf{Logical Sense }} & \multicolumn{3}{c|}{\textbf{Single Topic }} & \multicolumn{3}{c|}{\textbf{Enjoyable}} & \multicolumn{3}{c}{\textbf{Fluency}}&\textbf{Num}\\
    & \textbf{set} & \textbf{chars} & Win\% & Lose\% & Tie\% & Win\% & Lose\% & Tie\% &  Win\% & Lose\% & Tie\%  & Win\% & Lose\% & Tie\%&\textbf{story} \\
    \midrule

    \multirow{2}{*}{CAST vs \citeauthor{guan2020knowledge}} 
    &\multirow{2}{*}{ROC}
    & 1 & 
    \textbf{95}**&0&5&
    \textbf{90}**&5&5 & 
    \textbf{95}**&5&5&
    \textbf{95}**&5&5&20 \\
    
    && 2 &
    \textbf{90}**$\dagger$&5&5&
    \textbf{90}**&5&5&
    \textbf{95}**$\dagger$&5&0&
    \textbf{95}**$\dagger$&5&0&
    20
    \\
    
    \midrule
    CAST vs \citeauthor{goldfarb2020content}
    & WP
    & 1  &
    \textbf{60}$\ddagger$&30&10&
    \textbf{70}$\dagger$&30&0&
    \textbf{70}$\ddagger$&30&0&
    \textbf{60}$\dagger$&40&0&
    10 \\
    
    \midrule
    CAST vs C2PO
    &FT
    & 1 & 
    \textbf{90}**$\dagger$&10&0&
    \textbf{70}&20&10&
    \textbf{100}**&0&0&
    \textbf{100}*&0&0&
    10  \\

    \midrule
    \midrule
    \multirow{2}{*}{GPT-ROC-RL vs \citeauthor{guan2020knowledge}}
    &\multirow{2}{*}{ROC}
    & 1 & 
    \textbf{65}**$\dagger$&15&20&
    \textbf{55}$\dagger$&35&10&
    \textbf{50}$\dagger$&40&10&
    \textbf{50}&30&20&
    20\\
    
    && 2 &
    \textbf{60}$\dagger$&30&10&
    \textbf{50}&40&10&
    \textbf{65}&30&5&
    \textbf{70}*&15&15&
    20\\
    \midrule
    
    \multirow{2}{*}{GPT-ROC-RL vs CAST} &
    \multirow{2}{*}{ROC}
    &1 &
    15&\textbf{80}**$\dagger$&5&
    10&\textbf{80}**$\dagger$&10&
    20&\textbf{75}**$\dagger$&5&
    15&\textbf{65}**&20&
    20\\

    && 2 &
    5&\textbf{85}**$\dagger$&10&
    5&\textbf{80}**&15&
    100&\textbf{85}**$\dagger$&5&
    0&\textbf{95}**$\dagger$&5&
    20\\
    \bottomrule
\end{tabular}
\caption{Human-participant evaluation results for experiments 1 and 2, 
showing the percentage of participants who preferred the first system, second system, or thought the systems were equal.
Each system is conditioned on the same test-set prompts. * indicates results are significant at $p<0.05$ confidence level; ** at $p<0.01$ using a Wilcoxan sign test on win-lose pairs. 
$\dagger$ indicates $\kappa$ > 0.2 or fair agreement. $\ddagger$
indicates $\kappa$ > 0.4 or moderate
agreement.
}
\label{tab:CAST_results_majority}
\end{table*}

\section{Evaluation}
\label{app:eva}
\subsection{Evaluated Stories Generation}
\label{app:gen}
In order to compare with \citet{guan2020knowledge}, we randomly select a subset of first sentences of ROCStories as prompts to seed CAST and \citet{guan2020knowledge}, generating 5-sentence stories from each model. 
We considered two cases---(1) single-character and (2) two-character stories.
In order to generate two-character stories, we seed the story history and continuation's subject to GPT-2. More details can be found in Appendix~\ref{app:appendix-prefix}.
We use a subset of prompts given by Writing Prompts to seed our system and \citet{goldfarb2020content}.
We also keep 5 sentences to evaluate the models.
Since C2PO is controllable story generation model trained on fairy tale stories, we seed the first sentence in the fairy tale stories as prompt and the $5$th sentence in the story as goal to C2PO for generating stories. For CAST model, we only seed the first sentences of the fairy tale stories.
Examples can be found in Appendix~\ref{app:extra_stories}.

\subsection{Human Study Setup}
\label{app:human}
\begin{figure}[!tbh]
    \centering
    \includegraphics[width=\columnwidth]{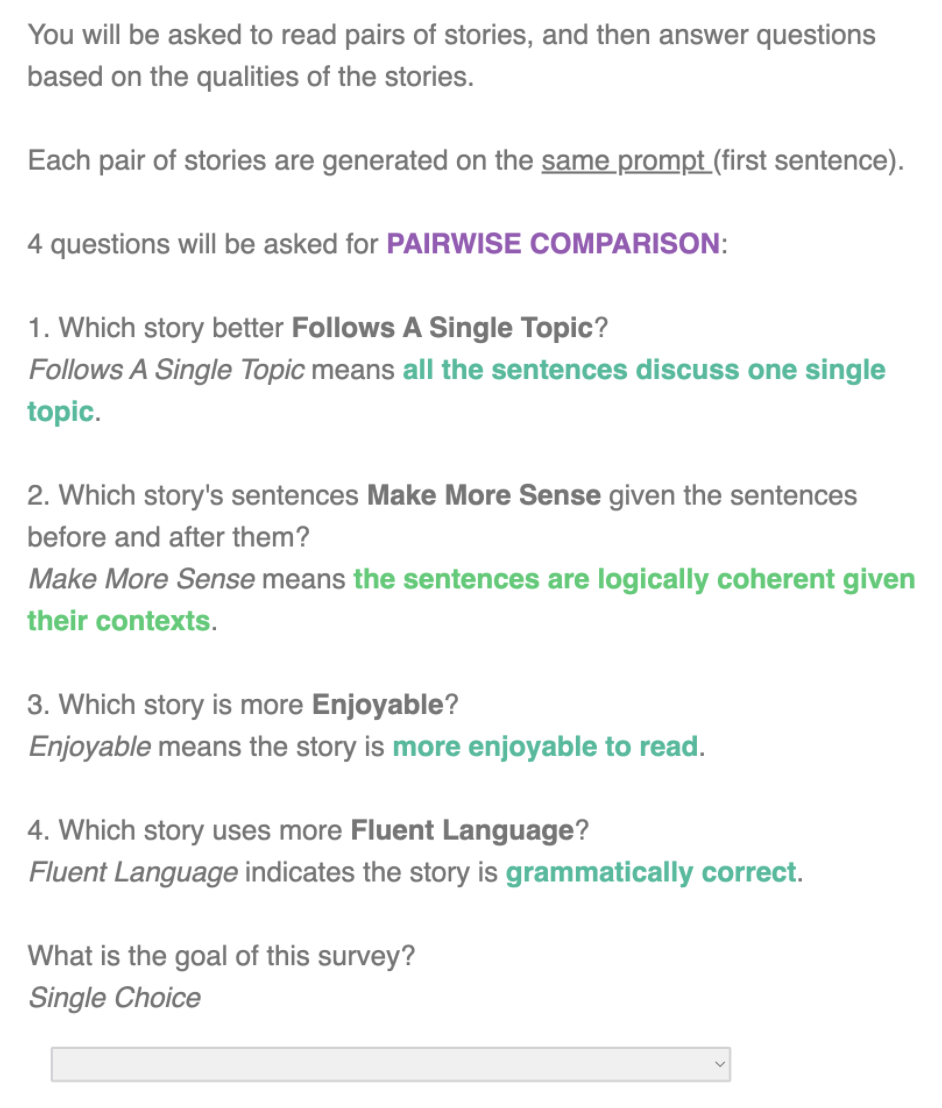}
    \caption{Instructions given to human study participants, along with a question to validate they have read them.}
    \label{fig:instr}
\end{figure}

\begin{figure}[!tbh]
    \centering
    \includegraphics[width=\columnwidth]{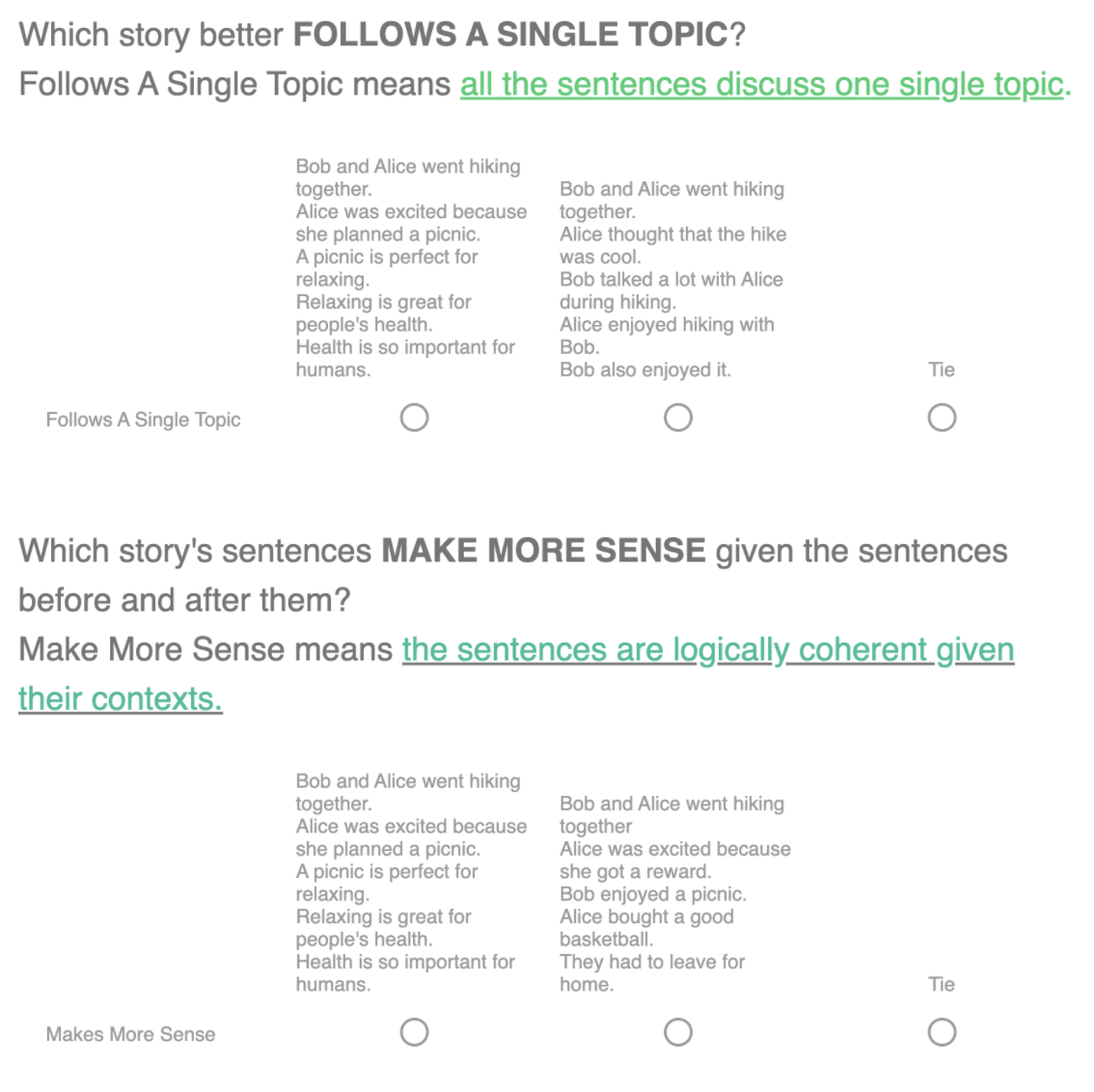}
    \includegraphics[width=\columnwidth]{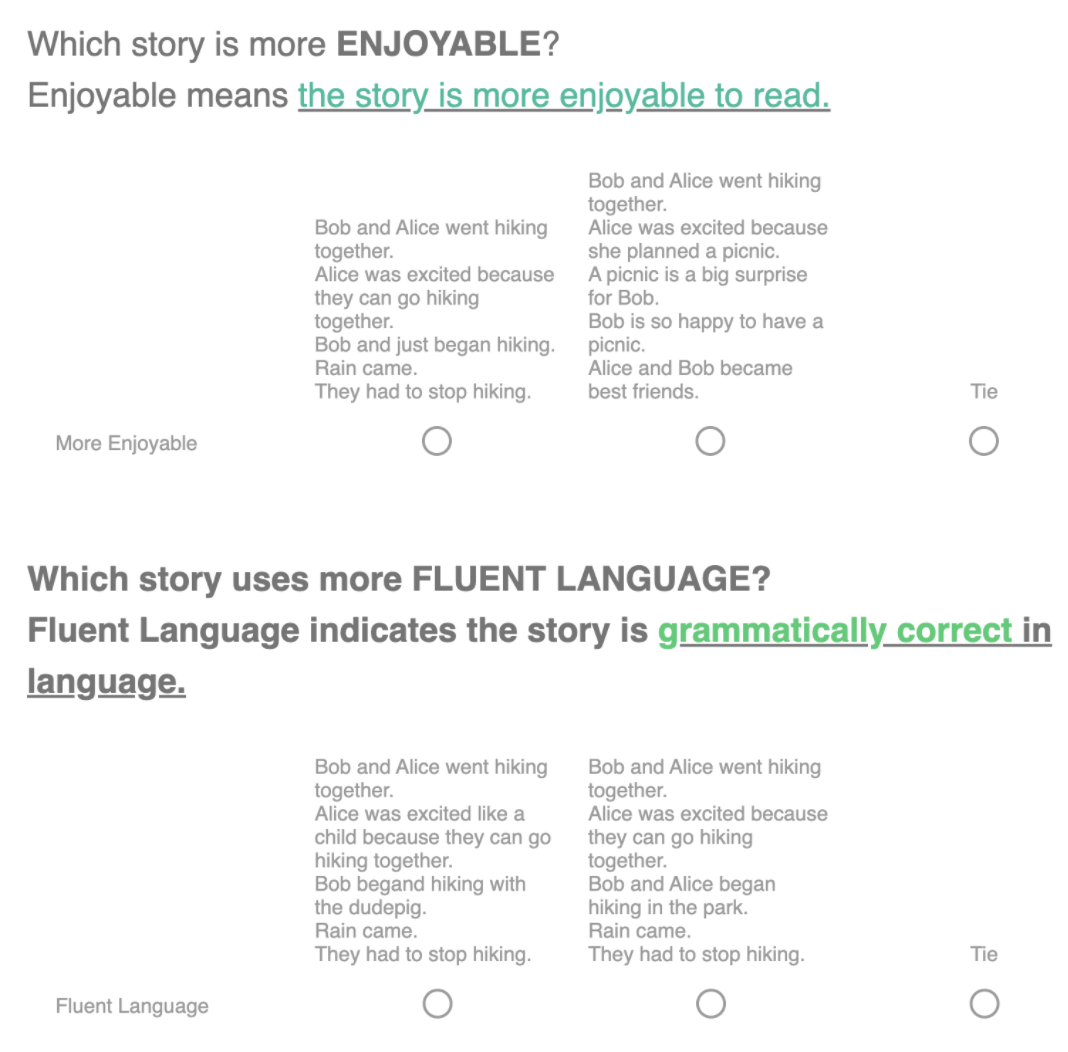}
    \caption{Screening questions used to qualify participants for the main study.
    Correct answers for these questions are [2,1,any,2], where 1 indicates the first story is the correct answer, 2 indicates the second one is the correct answer and ``any'' indicates no correct answers, any answer can pass the screen question.}
    \label{fig:s_1}
\end{figure}

\begin{figure}[!tbh]
    \centering
    \includegraphics[width=\columnwidth]{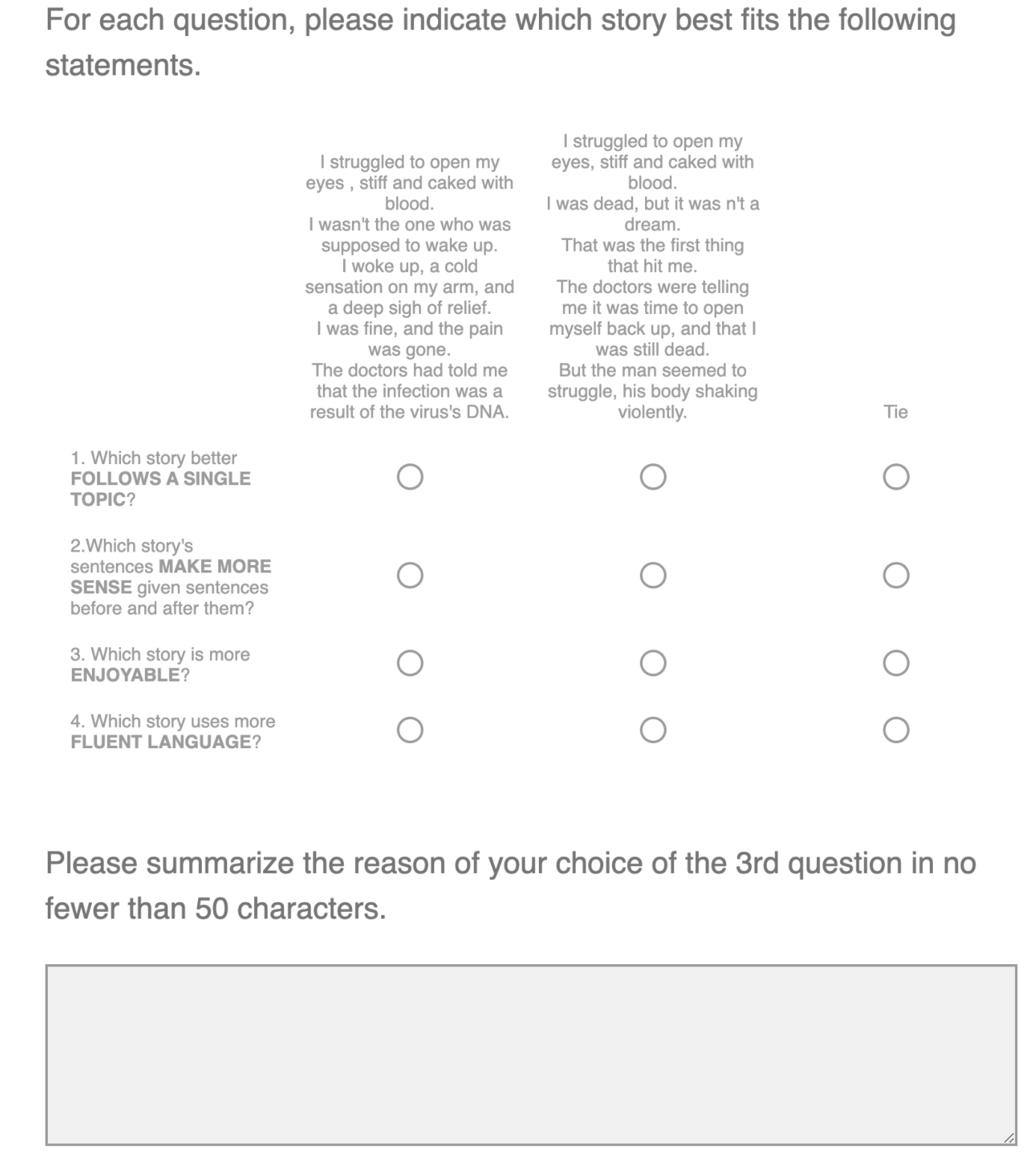}
    \caption{User interface for the main human evaluation task. Each participant completes 5 such tasks in a single HIT.}
    \label{fig:human_ex}
\end{figure}

We show human participants instructions
(Fig.~\ref{fig:instr}) and then they are required to pass screen questions (Fig.~\ref{fig:s_1}).
They then answer which story best met the criteria, as shown in Fig.~\ref{fig:human_ex}.

\begin{table*}
\begin{tabular}{|l|l|l|l|l|l|l|}
\hline
\textbf{Study}                   & \textbf{1 \& 2} & \textbf{1 \& 3} & \textbf{1 \& 4} & \textbf{2 \& 3} & \textbf{2 \& 4} & \textbf{3 \& 4} \\ \hline
CAST vs. C2PO                    & 0.39*          & 0.34           & 0.34           & 0.47*          & 0.77*          & 0.39*          \\ \hline
CAST vs. \citeauthor{goldfarb2020content}\shortcite{goldfarb2020content}                 & 0.58*          & 0.44*          & 0.38*          & 0.56*          & 0.35           & 0.47*          \\ \hline
CAST-ROC-1Char vs. \citeauthor{guan2020knowledge}\shortcite{guan2020knowledge}         & 0.51*          & 0.28*          & 0.43*          & 0.67*          & 0.65*          & 0.67*          \\ \hline
CAST-ROC-2Char vs. \citeauthor{guan2020knowledge}\shortcite{guan2020knowledge}         & 0.53*          & 0.49*          & 0.39*          & 0.91*          & 0.79*          & 0.79*          \\ \hline
CAST-RL-1Char vs. \citeauthor{guan2020knowledge}\shortcite{guan2020knowledge} & 0.78*          & 0.65*          & 0.47*          & 0.67*          & 0.54*          & 0.45*          \\ \hline
CAST-RL-2Char vs. \citeauthor{guan2020knowledge}\shortcite{guan2020knowledge}          & 0.51*          & 0.45*          & 0.44*          & 0.65*          & 0.65*          & 0.55           \\ \hline
CAST-RL-1Char vs. CAST-ROC-1Char & 0.83*          & 0.28*          & 0.55*          & 0.32*          & 0.64*          & 0.31*          \\ \hline
CAST-RL-2Char vs. CAST-ROC-2Char & 0.14           & -0.05          & 0.25           & 0.27           & 0.39*          & 0.30*          \\ \hline
\end{tabular}
\label{tab:spearman-nums}
\caption{Spearman correlation results from all our human subject studies. The four numbers 1, 2, 3, and 4 corresponds to ``Follow a single topic'', ``Logical Sense'', ``Enjoyable'', and ``Fluency'' metrics respectively. Spearman correlations with $p \leq 0.01$ are marked with an *.}
\end{table*}

\subsection{Correlation between answers}
\label{app:correlation}
We compute the Spearman rank correlations between the workers' different answers for the story pairs they are responsible for rating. We ignore workers who did not complete all of the questions in our computations. Our results are displayed in Table 11.

\clearpage
\section{Additional Story Outputs}
\label{app:extra_stories}
Additional 5-sentence story examples are as follows,

\setlength{\parindent}{0pt}
\rule{\linewidth}{1pt}

\textbf{Seed Prompt:}
    
Bob and Alice went hiking together. 

\rule{\linewidth}{1pt}

\textbf{CAST:}
    
Alice was excited because she planned a picnic. 

Bob was excited because he was staying up late at night.

Alice felt much better about her camping trip. 

Bob spent the night playing in the woods.

\rule{\linewidth}{0.5pt}

\textbf{CAST-RL:}

Alice decided to take a group photo. 

Bob shared the picture with Alice. 

Alice came over and smiled at Bob. 

Bob was so happy that he took the picture.

\rule{\linewidth}{0.5pt}

\textbf{\citeauthor{guan2020knowledge} \shortcite{guan2020knowledge}:}

Alice thought that the hike was cool. 

Bob was pretty cool. 

Alice reminded him of the great fall, especially in the forest. 

Bob was cute, but Alice was lost.

\rule{\linewidth}{1pt}
\rule{\linewidth}{1pt}

\textbf{Seed Prompt:}

Bob asked Alice to come over for dinner.

\rule{\linewidth}{1pt}

\textbf{CAST:}

Alice wanted to go to a party. 

Bob agreed to go with her. 

Alice and Bob went to the concert. 

Bob bought a bottle of wine, and they enjoyed their time.

\rule{\linewidth}{0.5pt}

\textbf{CAST-RL:}

Alice said yes, so he brought her food. 

Bob had dinner with Alice that night. 

Alice said yes to the dinner. 

Bob was happy that he invited her to dinner with him.

\rule{\linewidth}{0.5pt}

\textbf{\citeauthor{guan2020knowledge} \shortcite{guan2020knowledge}:}

Alice drove and came over Bob 's house in time. 

Bob decided to let Alice choose the foods. 

Alice ate some dog food and then she met her husband. 

Bob offered Alice a side of anything he had for dinner.

\rule{\linewidth}{1pt}
\rule{\linewidth}{1pt}

\textbf{Seed Prompt:}

Bob took Alice fishing in the summer.

\rule{\linewidth}{1pt}

\textbf{CAST:}

Alice waited at the lake for hours. 

Bob got another boat and brought it back home.

Alice and Bob fished the next day. 

Bob enjoyed the fishing and had a great time.

\rule{\linewidth}{0.5pt}

\textbf{CAST-RL:}

Alice caught ten fish. 

Bob was pleased that he caught ten fish. 

Alice was very pleased to have a good fish. 

Bob is glad that he caught ten fish.

\rule{\linewidth}{0.5pt}

\textbf{\citeauthor{guan2020knowledge} \shortcite{guan2020knowledge}:}

Alice got seasick. 

Bob asked Alice to help him. 

Alice went to Bob 's house to tow Bob 's boat. 

Bob got a surfboard for a big fishing trip.

\rule{\linewidth}{1pt}
\rule{\linewidth}{1pt}

\textbf{Seed Prompt:}

Bob was engaged to Alice. 

\rule{\linewidth}{1pt}

\textbf{CAST:}

Alice wanted a diamond ring. 

Bob thought that ring was the perfect diamond. 
Alice loved the diamond ring and decided to buy it. 

Bob went to the store and bought the ring.

\rule{\linewidth}{0.5pt}

\textbf{CAST-RL:}

Alice was going on a date with Bob. 

Bob agreed to the date and married her. 

Alice now has two kids and needs the baby. 

Bob is very happy he finally realized she was unhappy.

\rule{\linewidth}{0.5pt}

\textbf{\citeauthor{guan2020knowledge} \shortcite{guan2020knowledge}:}

Alice just got a new husband. 

Bob wondered what to expect. 

Alice bought the nice wedding ring from a daycare. 

Bob wore it to the wedding.

\rule{\linewidth}{1pt}
\rule{\linewidth}{1pt}

\textbf{Seed Prompt:}

Bob took Alice to school today. 

\rule{\linewidth}{1pt}

\textbf{CAST:}

Alice was a freshman. 

Bob took her to the library. 

Alice was excited to take the long bus ride to the library. 

Bob continued the long walk through the library.

\rule{\linewidth}{0.5pt}

\textbf{CAST-RL:}

Alice had been waiting for hours at school. 

Bob was relieved to see her take the bus to school. 

Alice was excited to see the bus arrive.

Bob was proud of her for taking the bus to school.

\rule{\linewidth}{0.5pt}

\textbf{\citeauthor{guan2020knowledge} \shortcite{guan2020knowledge}:}

Alice was nervous because Bob didn't know how to tap dance. 

Bob was excited for his first attempt at tap dancing. 

Alice was impressed and wanted to learn. 

Bob started his first dance at the dance school.

\rule{\linewidth}{1pt}

Additional 10-sentence story examples are as follows:

\rule{\linewidth}{1pt}

\textbf{Seed Prompt:}

Bob went on her very first date with Alice.

\rule{\linewidth}{1pt}

\textbf{CAST:}

Alice was very excited about having a date with Bob. 

Bob was excited and eager. 

Alice informed Bob she was very excited for him. 

Bob was very appreciative. 

Alice said it was a perfect time. 

Bob and Alice had a wonderful first date.

Alice was surprised she was invited to his birthday. 

Bob was thrilled. 

Alice was happy to have a new boyfriend.

\rule{\linewidth}{0.5pt}

\textbf{\citeauthor{guan2020knowledge} \shortcite{guan2020knowledge}:}

Alice and Bob didn't realize that Bob loved movies. 

Bob's friend, Alice, saw the movie with Alice. 

Alice and Bob were good friends. 

Bob's friend, Alice, screamed for 
Alice. 

Alice was their eyes met her eyes. 

Bob and Alice's grandparents were never match.

Alice because they had ever though, they thought the end the words said, she passed with Alice. 

Bob and they agreed that they went out the eyes. 

Alice felt they got puppies and they do the thought she was their eyes Alice had.

\rule{\linewidth}{1pt}

\rule{\linewidth}{1pt}

\textbf{Seed Prompt:}

Bob broke up with Alice. 

\rule{\linewidth}{1pt}

\textbf{CAST:}

Alice was very hurt and very angry. 

Bob knew he would have to live with his pain. 

Alice said she could give him a relief. 

Bob felt much better after that. 

Alice felt even better in peace. 

Bob is now happy with himself. 

Alice felt much better. 

Bob will make himself a happy man. 

Alice feels much better after having peace.

\rule{\linewidth}{0.5pt}

\textbf{\citeauthor{guan2020knowledge} \shortcite{guan2020knowledge}:}

Alice didn't want to get it. 

Bob thought that he was too thin. 

Alice said he could give back inside a horse. 

Bob could bring along for a ride and to return to the river. 

Alice had to drive for three days and she realized she was cheating. 

Bob and Alice got into a fight. 

Alice ran away with Alice and grounded.

Bob and sent Alice. 

Alice and a girl kicked each other.

\rule{\linewidth}{1pt}

\rule{\linewidth}{1pt}

\textbf{Seed Prompt:}

Bob went to see Alice today. 

\rule{\linewidth}{1pt}

\textbf{CAST:}

Alice told him about the movie she loved to watch. 

Bob grabbed a movie poster from the library. 

Alice shared her own reaction to the movie.

Bob watched the movie that night. 

Alice felt very satisfied after the movie was over. 

Bob went home happy that he watched the movie.

Alice has been a huge fan of the movie ever since. 

Bob looks forward to a nice vacation at the movies. 

Alice is glad that she shared her own comments to the movie.

\rule{\linewidth}{0.5pt}

\textbf{\citeauthor{guan2020knowledge} \shortcite{guan2020knowledge}:}

Alice was nice and gregarious. 

Bob had a great time. 

Alice had some great food and wine. 

Bob couldn't wait to see Alice again.

Alice let him take a bath. 

Bob felt smarter and had a good time again.

Alice also got his great seeing Alice tattoo. 

Bob had a good time. 

Alice tattooy from his favorite j tomorrow too.

\rule{\linewidth}{1pt}

\rule{\linewidth}{1pt}

\textbf{Seed Prompt:}

Bob went over to play at Alice's house.

\rule{\linewidth}{1pt}

\textbf{CAST:}

Alice liked to watch movies. 

Bob was always there watching movies. 

Alice saw a movie she wanted to see. 

Bob was very glad to buy that movie. 

Alice was so happy with the movie he bought.

Bob was excited to be watching the movie with Alice. 

Alice could not wait to see the movie with Bob. 

Bob and Alice had so many fun. 

Alice laughed and thanked Bob for the great movie.

\rule{\linewidth}{0.5pt}

\textbf{\citeauthor{guan2020knowledge} \shortcite{guan2020knowledge}:}

Alice was a baby dolls. 

Bob told Alice he could save a doll.

Alice went to the video game section to change the game. 

Bob was amazed to see who did the laundry to the last minute. 

Alice waved and proposed the doll girls.

Bob wished her that she was getting a doll.

Alice helped him with the \$40 and she gave her \$100, but by buttering buttering a doll. 

Bob's doll broke.

Alice had an ruined doll and they would buy her.

\rule{\linewidth}{1pt}

\end{document}